%% file: main.tex
\documentclass[dvipsnames]{article}
\usepackage[preprint]{colm2025_conference}
\vspace{0pt} 
\usepackage{parskip}
\input{math_commands.tex}

\usepackage{hyperref}
\usepackage{url}
\usepackage{booktabs}
\usepackage{multirow}
\usepackage{graphicx}
\usepackage{float}
\usepackage{titlesec}
\usepackage{latexsym}
\usepackage{caption}
\usepackage{colortbl}
\usepackage[most]{tcolorbox}
\usepackage{subcaption}
\usepackage{helvet}
\usepackage{xfrac}
\usepackage{algorithm}
\usepackage{algpseudocode}
\usepackage{wrapfig}
\usepackage{bbm}
\usepackage{amsmath}   % for \dfrac, \ell, alignments, …
\usepackage{amssymb}   % (optional) extra math symbols
\usepackage{enumitem}
\usetikzlibrary{calc}
\usepackage{yhmath}

\definecolor{grpo}{HTML}{FAD4D4}   % very light red/pink tone
\definecolor{gfpo}{HTML}{BEE4D0}   % cool green
\definecolor{comparison}{HTML}{B3D9FF} % light blue

\definecolor{grayrowcolor}{RGB}{220,220,220} % Light gray color

\definecolor{commentcolour}{rgb}{0.3,0.7,0.2}
\definecolor{lightblue}{RGB}{245, 250, 250} % lighter & softer
\definecolor{blue}{RGB}{77, 174, 172} % softer teal

\definecolor{darkpink}{RGB}{255, 105, 180}

\def\shownotes{0}  %set 1 to show author notes
\ifnum\shownotes=1
\newcommand{\authnote}[2]{[#1: #2]}
\else
\newcommand{\authnote}[2]{}
\fi

\newtcolorbox{findings}{
    enhanced,
    breakable,
    colback=lightblue,         %
    colframe=blue,        %
    boxrule=1.5pt,
    arc=0.25em,
    left=1em,
    right=1em,
    top=1em,
    bottom=0.75em,
    before=\vspace{1em},
    overlay unbroken and first={
        \node[
            fill=blue,
            text=white,
            font=\bfseries,
            anchor=west,
            inner xsep=0.75em,
            inner ysep=0.5em,
            rounded corners=0.25em
        ] 
        at ([xshift=0.75em]frame.north west) {Finding};
    }
}

\tcbset{
  aibox/.style={
    width=\textwidth,
    top=0pt, bottom=0pt, left=5pt, right=5pt,
    colback=white,
    colframe=black,
    colbacktitle=black,
    enhanced,
    center,
    attach boxed title to top left={yshift=-0.1in,xshift=0.15in},
    boxed title style={boxrule=0pt,colframe=white,},
  }
}
\newtcolorbox{AIbox}[2][]{aibox,title=#2,#1}
\definecolor{user}{HTML}{F2EFE7}  % for the user box
\definecolor{model}{HTML}{81E7AF} %81E7AF} % for the model box

\newcounter{mySubSec}[section] % resets each section
\renewcommand{\themySubSec}{\thesection.\arabic{mySubSec}}

\newcommand{\mySubsection}[1]{%
    \refstepcounter{mySubSec}% makes label/refs work
    \phantomsection
    \noindent\textbf{\themySubSec\hspace{1ex} #1}%
}

\title{Sample More to Think Less: \\Group Filtered Policy Optimization for Concise Reasoning}

\newcommand{\authorsep}{\hspace{2ex}}
\newcommand{\instsep}{\hspace{2ex}}

 {\author{Vaishnavi Shrivastava$^{m}$ \authorsep 
Ahmed Awadallah$^m$ \authorsep Vidhisha Balachandran$^m$\\ 
\textbf{Shivam Garg$^m$ \authorsep Harkirat Behl$^m$ \authorsep Dimitris Papailiopoulos$^{m,\omega}$}\\\\
$^{m}$Microsoft Research \instsep\\
$^{\omega}$University of Wisconsin-Madison
}}

\begin{document}

\maketitle

\input{sections/abstract}

\input{sections/introduction}
\input{sections/preliminaries}
\input{sections/method}
\input{sections/setup}
\input{sections/results}
\input{sections/analysis}
\input{sections/related_work}
\input{sections/conclusion}
\input{sections/acknowledgments}
\newpage
\bibliography{colm2025_conference}
\bibliographystyle{colm2025_conference}
\newpage
\input{sections/appendix}
\end{document}

%% file: math_commands.tex
\usepackage{amsmath,amsfonts,bm}

\def\eqref#1{equation~\ref{#1}}
\def\1{\bm{1}}

\DeclareMathAlphabet{\mathsfit}{\encodingdefault}{\sfdefault}{m}{sl}
\SetMathAlphabet{\mathsfit}{bold}{\encodingdefault}{\sfdefault}{bx}{n}

%% file: sections/abstract.tex
\begin{abstract}
Large language models trained with reinforcement learning with verifiable rewards tend to trade accuracy for length—inflating response lengths to achieve gains in accuracy. While longer answers may be warranted for harder problems, many tokens are merely “filler”: repetitive, verbose text that makes no real progress. We introduce GFPO (\textit{\textbf{G}roup \textbf{F}iltered \textbf{P}olicy \textbf{O}ptimization}), which curbs this length explosion by sampling larger groups per problem during training and filtering responses to train on based on two key metrics: (1) response length and (2) token efficiency: reward per token ratio. By sampling \textit{more} at training-time, we teach models to think \textit{less} at inference-time. On the Phi-4-reasoning model, GFPO cuts GRPO’s length inflation by 46–71\% across challenging STEM and coding benchmarks (AIME 24/25, GPQA, Omni-MATH, LiveCodeBench) while maintaining accuracy. Optimizing for reward per token further increases reductions in length inflation to 71–85\%. We also propose Adaptive Difficulty GFPO, which dynamically allocates more training resources to harder problems based on real-time difficulty estimates, improving the balance between computational efficiency and accuracy especially on difficult questions. GFPO demonstrates that increased training-time compute directly translates to reduced test-time compute---a simple yet effective trade-off for efficient reasoning.
\end{abstract}

%% file: sections/introduction.tex
\input{figures/intro_fig}
\section{Introduction}
Reinforcement learning from verifier rewards (RLVR) methods such as GRPO \citep{grpo} and PPO \citep{schulman2017proximal} have been pivotal in enabling test-time scaling—allowing models like O3 \citep{openai2025o3o4mini} and DeepSeek-R1 \citep{deepseekr1} to ``think longer’’ and unlock unprecedented performance on challenging reasoning tasks such as AIME and IMO. While longer reasoning chains are expected for solving harder problems, prior work shows that length inflation can be uncorrelated with correctness, and that \textit{shorter} chains may in fact yield \textit{better} accuracy. For example, \cite{balachandran2025inference} report that on AIME 25, DeepSeek-R1 generates responses nearly 5x longer than Claude 3.7 Sonnet, despite achieving similar accuracy. Likewise, \cite{hassid2025don} find that on AIME and HMMT, the shortest responses from QwQ-32B outperform randomly sampled responses by 2\% while using 31\% fewer tokens, indicating that longer chains are not synonymous with better reasoning.

Longer responses can appear less accurate simply because they often arise from harder questions. To disentangle genuine length increases driven by question difficulty from unnecessary inflation, we analyze the correlation between response length and correctness for multiple responses to the \textit{same} questions in Phi-4-reasoning-plus \citep{abdin2025phi4reasoningtechnicalreport}. On AIME 25, we find that in 72\% of questions where both correct and incorrect responses are generated, longer responses are more likely to be wrong than their shorter counterparts.

Approaches such as Dr. GRPO \citep{drgrpo} and DAPO’s \citep{dapo} token-level loss normalization have been proposed to curb the persistent length inflation phenomenon in RLVR-trained models. Yet, even with token-level normalization applied during the training of Phi-4-reasoning-plus, we observe rapid response length growth—from \textit{4k} to \textit{14k} tokens in just 100 steps of GRPO training. We hypothesize that while token-level normalization penalizes long incorrect responses more heavily, it also amplifies rewards for long correct chains—unintentionally reinforcing the inherent verbosity of strong base models that have been heavily SFTed for step-by-step reasoning (e.g., Phi-4-reasoning \citep{abdin2025phi4reasoningtechnicalreport} and DeepSeek-R1-Distill-Qwen \citep{deepseekr1}). This underscores the difficulty of relying on loss normalization alone to counteract GRPO’s pronounced length inflation.

Motivated by these observations, our goal is to develop \textit{efficient} reasoning models—models that retain the reasoning accuracy afforded by GRPO while producing substantially shorter reasoning chains.
Towards achieving this goal, we make the following contributions:
\begin{itemize}[leftmargin=3em]
\item \textbf{GFPO (Group Filtered Policy Optimization)}: We propose GFPO (Figure~\ref{fig:overview}, Section~\ref{sec:gfpo}), a simple yet effective variant of GRPO designed to explicitly counteract response length inflation. GFPO combines rejection sampling with standard GRPO: for each question, we sample a larger group of candidate reasoning chains $G$ to increase exposure to desirable outputs, filter them according to a target metric, and only learn from the policy gradients of the top-$k$ retained chains. While many rejection metrics are possible, we focus on response length—retaining the shortest chains to encourage the model to “think less” while reasoning.

When optimized for length, GFPO reduces GRPO's length inflation by 46.1\% on AIME 25, 59.8\% on AIME 24, 57.3\% on GPQA, 71\% on Omni-MATH, and 57\% on LiveCodeBench, all while maintaining accuracy (Section~\ref{res:grpo_efficient},~\ref{res:vary_k_G}).
\item \textbf{Token Efficiency} (Section~\ref{res:tok_eff}): Beyond targeting length alone, we introduce the \textit{token efficiency} metric—defined as the ratio of reward to response length. 
This metric promotes reasoning chains that justify their length by delivering proportionally higher rewards, encouraging the model to be both concise and effective.

Optimizing for token efficiency with GFPO reduces length inflation by 70.9\% on AIME 25, 84.6\% on AIME 24, 79.7\% on GPQA, 82.6\% on Omni-MATH, and 79.7\% on LiveCodeBench (see Appendix~\ref{appendix:qual_ex} for qualitative examples).
\item \textbf{Adaptive Difficulty GFPO} (Section~\ref{res:adaptive_diff}): We further introduce an adaptive variant of GFPO in which the number of retained responses $k$ is dynamically adjusted based on a lightweight, unsupervised estimate of question difficulty. This adaptive strategy allocates more exploration (larger $k$) to harder questions, while aggressively shortening easier ones.
\begin{figure}[t]
    \centering
    \begin{subfigure}[b]{0.49\textwidth}
        \includegraphics[width=\textwidth]{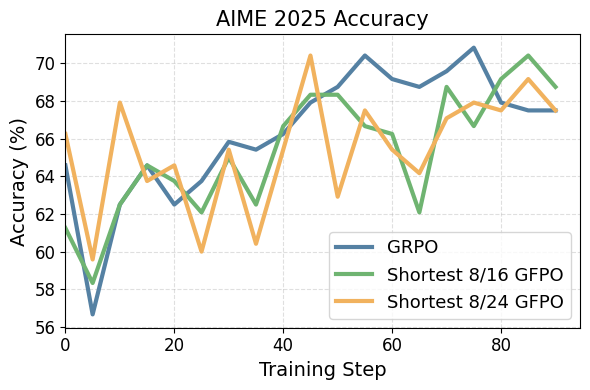}   
    \end{subfigure}
    \hfill
    \begin{subfigure}[b]{0.49\textwidth}
        \includegraphics[width=\textwidth]{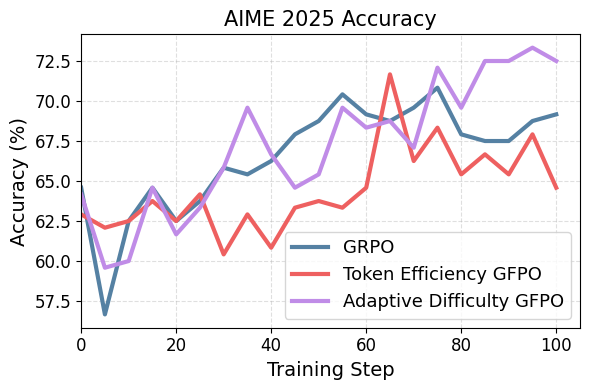}
    \end{subfigure}
    \caption{\textbf{Comparison of GFPO and GRPO on AIME 25} accuracy ($n=8$ samples) during training. GFPO variants reach the same peak performance as GRPO (blue).}
    \label{fig:aime_training}
\end{figure}
\item \textbf{Out-of-Distribution Generalization} (Section~\ref{res:ood}): We show that GFPO preserves accuracy while curbing length inflation even on out-of-distribution tasks.
\item \textbf{Analysis of GFPO on Response Length and Question Difficulty} (Section~\ref{sec:analysis}): Finally, we present a detailed analysis of GFPO’s accuracy and length reductions on easy vs. hard questions, and examine its impact on the accuracy of long responses.
\end{itemize}

GFPO exploits a fundamental trade-off between training and inference-time compute, shifting cost from inference—where shorter chains deliver substantial efficiency—to training, by sampling and evaluating additional candidate responses. This trade-off is particularly advantageous because training compute is a one-time investment, whereas inference compute savings are realized continuously throughout deployment.
In doing so, GFPO offers a simple yet effective solution to the response-length inflation inherent in reasoning models—retaining GRPO’s state-of-the-art performance while producing dramatically shorter reasoning chains.

%% file: figures/intro_fig.tex
\begin{figure}[H]
  \centering

  % ----- LEFT: stacked PDFs (0.55 of line width) -----
  \begin{subfigure}[b]{0.635\linewidth}
    \includegraphics[width=\linewidth]{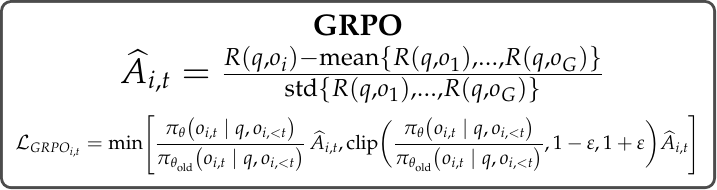}\\[6pt] % small negative to tighten
    \includegraphics[width=\linewidth]{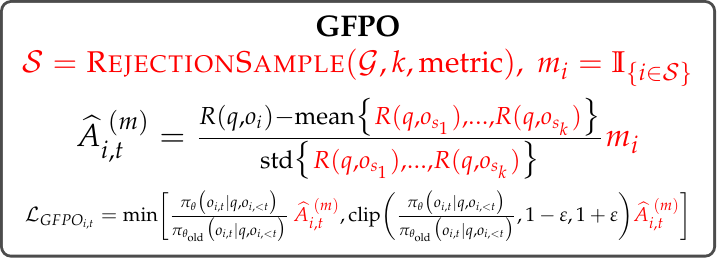}
  \end{subfigure}%
  \hfill
  % <-- no \hfill, no space; % keeps subfigures flush
  % ----- RIGHT: plot (0.45 so the sum is 1.00) -----
  \begin{subfigure}[b]{0.35\linewidth}
    \includegraphics[width=\linewidth]{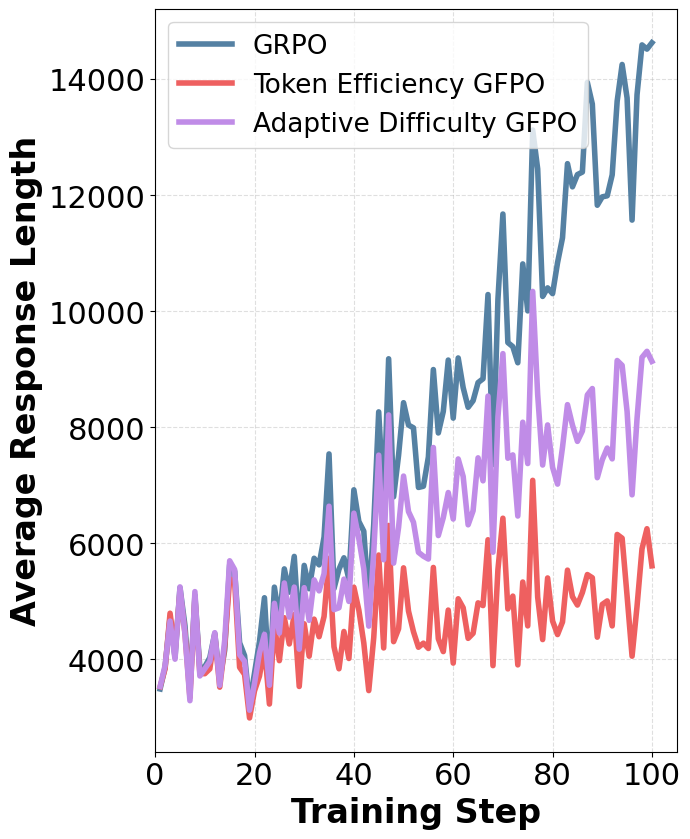}
  \end{subfigure}
    \caption{\textbf{Left:} GFPO introduces simple yet powerful modifications to GRPO: sample more responses during training ($\uparrow G$), rank them by a target attribute (e.g., length, token efficiency), and learn only from the top-$k$—setting the advantages of the rest to zero. This selective learning functions as implicit reward shaping, steering the policy toward desired behaviors.
\textbf{Right:} When optimizing for length or token efficiency, GFPO curbs GRPO’s length inflation—letting the model \textit{think less} at inference-time by \textit{sampling more} at training-time—while maintaining its core reasoning capabilities.}
  \label{fig:overview}
\end{figure}

%% file: sections/preliminaries.tex
\section{Preliminaries}
Group Relative Policy Optimization (GRPO; \cite{grpo}) is a reinforcement learning algorithm that simplifies Proximal Policy Optimization (PPO; \cite{schulman2017proximal}) by eliminating the need for a value model to estimate the baseline advantage. 
This is achieved by sampling multiple responses per question and using their average reward as a baseline, while still optimizing a similar clipped surrogate objective as PPO.
Let $\theta$ denote the model parameters, $q$ denote the question, and $o$ denote responses sampled from the old policy $\pi_{\theta_\text{old}}$. The GRPO objective can then be written as:

\vspace{-0.5\baselineskip}
\begin{align}
\mathcal{J}_{\text{GRPO}}(\theta)
  &= \mathbb{E}_{[\substack{
        q \sim P(Q) ,\;
        \{o_i\}_{i=1}^{G} \sim \pi_{\theta_{\text{old}}}(O \mid q)
      }]}
     \frac{1}{G}
     \sum_{i=1}^{G}\frac{1}{ |o_i|}\sum_{t=1}^{|o_i|}
     \min\!\Bigl(
       r_{i,t}\widehat{A}_{i,t},
       \operatorname{clip}(r_{i,t},1-\varepsilon,1+\varepsilon)\widehat{A}_{i,t}
     \Bigr)                                   \notag \\
  &\quad
    -\beta\,\mathcal{D}_{KL}\!\bigl(\pi_\theta\!\parallel\!\pi_{\theta_{\mathrm{old}}}\bigr)
    +\gamma\,\mathrm{Entropy}(\pi_\theta)      \label{eq:grpo}
\end{align}
\\
where the advantage is $\widehat{A}_{i,t}=
\frac{R(q,o_i)-{\frac{1}{k}{\sum_{j=1}^{G}} R(q,o_j)}}{{{\sqrt{\frac{1}{k}\sum_{l=1}^{G}\bigl(R(q,o_i)-{{\frac{1}{k}{\sum_{j=1}^{G}} R(q,o_j)}}\bigr)^2}}}}$, 
$r_{i,t}=
\frac{\pi_\theta(o_{i,t}\,|\,q,o_{i,<t})}
     {\pi_{\theta_{\mathrm{old}}}(o_{i,t}\,|\,q,o_{i,<t})}$ and  $\beta\,\mathcal{D}_{KL}(\pi_\theta\!\parallel\!\pi_{\theta_{\mathrm{old}}})$ denotes the KL penalty.
     
Note that although we show the standard GRPO loss normalization equation, several open-source RL libraries including \texttt{verl}~\citep{sheng2024hybridflow} and \texttt{TRL}~\citep{vonwerra2022trl} default to the DAPO token-level loss normalization for GRPO, which is also what we use in our experiments.

A key limitation of GRPO is its reliance on a single scalar reward signal, making it difficult to jointly optimize multiple desirable response attributes, such as brevity and accuracy. This often leads to gains in accuracy at the cost of substantial response length inflation. To address this, we introduce GFPO to enable simultaneous optimization of multiple response properties.

%% file: sections/method.tex
\section{Group Filtered Policy Optimization}\label{sec:gfpo}
We propose \textbf{G}roup \textbf{F}iltered \textbf{P}olicy \textbf{O}ptimization (GFPO), a simple yet effective method for targeted policy optimization of desirable response properties. 
GFPO samples a larger group of candidate responses per question, broadening the response pool to include more candidates with desirable traits, and then explicitly filters for these traits when computing the policy gradient. 
While it may seem natural to directly encode desirable attributes such as brevity or informativeness into the scalar reward, doing so for multiple traits can be challenging, especially when correctness must already be captured.

Data filtration instead serves as an implicit, flexible form of reward shaping—akin to iterative self-improvement methods that use selective sampling to amplify specific model behaviors~\citep{zelikman2022star}. After this explicit filtering step isolates the preferred responses, standard rewards are then used solely to compute relative advantages within the selected group. Thus, GFPO optimizes for multiple desirable properties (e.g., length and accuracy) simultaneously, without requiring complex reward engineering. Since our goal is to reduce the response length inflation in RL, we focus on using GFPO to optimize for shorter responses while matching GRPO's accuracy.

\input{algorithms/rs}
Given a question $q$, we sample a large set of responses $\mathcal{G} = \{o_1, \ldots, o_G\}$ from the current policy. Rather than training equally on all responses, GFPO applies a selection step based on a user-specified metric to filter a subset of size $k$ of the most desirable responses to train on. We compute a metric score for each response and sort accordingly, selecting the top-$k$ responses to form the retained subset $\mathcal{S} \subseteq \mathcal{G}$ (Algorithm~\ref{alg:rej_sample}). We define a binary mask $m \in \{0,1\}^{G}$, where $m_i=1$ indicates a selected response and $m_i=0$ indicates a rejected response.

Formally, we define the GFPO objective\footnote{Note we use the DAPO token-level loss aggregation for both GFPO and GRPO which is the default choice in \texttt{verl}. We employ a slightly modified version of the clipped surrogate policy gradient loss introduced in prior work~\citep{li2025minimax}, which reduces training instabilities caused by negative advantages and large policy ratios.} as:
\begin{align}
\mathcal{J}_{\text{GFPO}}(\theta)
  &= \mathbb{E}_{\substack{
        q \sim P(Q),\;
        \{o_i\}_{i=1}^{G} \sim \pi_{\theta_{\text{old}}}(O \mid q)
      }}
     \frac{1}{\sum_{i=1}^{G}|o_i|}
     \sum_{i=1}^{G}\sum_{t=1}^{|o_i|}
     \min\!\Bigl(
       r_{i,t}\,{\color{Red}{\widehat{A}_{i,t}^{\;(m)}}},
       \operatorname{clip}(r_{i,t},1-\varepsilon,1+\varepsilon)\,
       {\color{Red}{\widehat{A}_{i,t}^{\;(m)}}}
     \Bigr)                                \notag \\[6pt]
% ─── centred second row ─────────────────────────────────────────────────────
\multicolumn{2}{c}{
  $-\beta\,\mathcal{D}_{KL}\!\bigl(\pi_\theta\!\parallel\!\pi_{\theta_{\mathrm{old}}}\bigr)
  +\gamma\,\mathrm{Entropy}(\pi_\theta)$
}                                          \label{eq:gfpo}
\end{align}
where
\[
{\color{red}{\mathcal{S}, m = \textsc{RejectionSample}(\mathcal{G}, k, \text{metric}, \text{order}),}}\;
      \color{red}{m_i=
      \mathbb{I}_{\{i\in \mathcal{S}\}}}
\]
\vspace{-0.1pt}
\[
\mu_S \;=\; \frac{1}{k}{\sum_{\color{Red}i\in S}} R(q,o_i),
\qquad
\sigma_S \;=\;
\sqrt{\frac{1}{k}\sum_{{\color{Red}i\in S}}\bigl(R(q,o_i)-{\color{Red}{\mu_S}}\bigr)^2},
\qquad
\widehat{A}_{i,t}^{\;(m)} \;=\;
 \frac{R(q,o_i)-{\color{Red}{\mu_S}}}{{\color{Red}{\sigma_S}}}{\color{Red}{m_i}}.
\]
\[
r_{i,t} \;=\;
\frac{\pi_\theta(o_{i,t}\,|\,q,o_{i,<t})}
     {\pi_{\theta_{\mathrm{old}}}(o_{i,t}\,|\,q,o_{i,<t})}.
\]

We normalize the advantages for responses in the selected subset $\mathcal{S}$ using the mean ($\mu_{S}$) and standard deviation ($\sigma_{S}$) of the response-level rewards in $\mathcal{S}$. This enables meaningful comparisons among responses already exhibiting the desired property, ensuring GFPO prioritizes the highest-reward responses within the filtered subset. Responses not in $\mathcal{S}$ receive zero advantage, effectively excluding them from influencing policy updates. Thus, GFPO’s primary intervention is at the level of advantage estimation, making it compatible with any GRPO variant such as DAPO \citep{dapo}, Dr. GRPO \citep{drgrpo}, or GRPO with the Dual-Clip PPO loss \citep{ye2020mastering}. Although GFPO incurs higher training-time compute by sampling more responses, this cost is partially offset as the learned policy produces shorter responses than GRPO.

While GFPO is general-purpose and can accommodate various scoring metrics, our experiments specifically leverage metrics aimed at reducing response length inflation:
\begin{itemize}
    \item \textbf{Response Length}: Training on short responses directly encourages brevity.
    \item \textbf{Token Efficiency} ($\text{reward} / \text{length}$): Training on highly token-efficient responses encourages succinctness, but still allows longer responses if sufficiently ``justified'' by proportionately higher rewards.
\end{itemize}
Other metrics—such as factuality, diversity, or external quality scores—could also be integrated into GFPO to optimize different attributes of interest.

\mySubsection{\textbf{Adaptive Difficulty GFPO.}}\label{sec:adaptive_diff_method}
We also introduce an adaptive difficulty variant of GFPO (Algorithm~\ref{alg:adaptive_gfpo}), aiming to allocate more training signal to harder questions. \input{algorithms/adaptive_diff}At each step of training, we estimate question difficulty by computing the average reward of sampled responses per question—lower average rewards indicate higher difficulty.

To adaptively scale the number of retained responses ($k$), we maintain a streaming summary of prompt difficulties using a lightweight t-digest data structure. The t-digest efficiently approximates quartiles over all prompt difficulties (reward means) seen thus far, enabling us to categorize new questions into relative difficulty buckets. Based on this categorization, we assign each question a target number of retained responses $k$: 4 for easy, 6 for medium, and 8 for hard and very hard questions (out of 16 sampled).\footnote{To prevent biased difficulty estimates due to insufficient processed prompts, for a few warmup training steps we retain 8 responses for all questions.} This dynamic curriculum enables more aggressive filtering on easy prompts, and more exploration on hard ones. The number of difficulty buckets and the $k$ per bucket are hyperparameters of this approach. 

Adaptive Difficulty GFPO makes efficient use of training compute, focusing gradient updates where they are most needed. It helps the model reduce verbosity on easy examples—where correctness is already high—while preserving accuracy on harder prompts by retaining more reasoning chains. To the best of our knowledge, this is the first algorithm to dynamically adapt the effective group size based on question difficulty.

%% file: algorithms/rs.tex
\begin{wrapfigure}{r}{0.55\textwidth}   % r/l = side,  width of the box
  \vspace{-2.1\baselineskip}            % nudge upward if needed
  \begin{minipage}{\linewidth}          % non-floating wrapper
    \begin{algorithm}[H]               % [H] keeps it HERE
      \caption{\textsc{Rejection Sampling}}
      \label{alg:rej_sample}
      \begin{algorithmic}[1]
        \Require group of responses $\mathcal{G}=\{o_1,\dots,o_G\}$,
                retain count $k$ $(k<G)$,
                scoring function $\mathit{metric}(\cdot)$,
                sort $\mathit{order}$ ($\uparrow$/$\downarrow$)
        \State $\mathbf{scores}\gets[\mathit{metric}(o_i)]_{i=1}^{G}$
        \State $\mathbf{idx}\gets\Call{argsort}{\mathbf{scores},\mathit{order}}$
        \State $\mathcal{S}\gets[\mathbf{idx}[j]\text{ for }j=1,\dots,k]$
        \State $m\gets[\mathbb{I}[\,i\in \mathcal{S}\,]\text{ for }i=1,\dots,G]$
        \State \Return $\mathcal{S},\;m$
      \end{algorithmic}
    \end{algorithm}
  \end{minipage}
  \vspace{-0.8\baselineskip}            % nudge the text gap below
\end{wrapfigure}

%% file: algorithms/adaptive_diff.tex
\begin{wrapfigure}{r}{0.60\textwidth}
  \vspace{-1.9\baselineskip}
  \begin{minipage}{\linewidth}
  \begin{algorithm}[H]
    \caption{\textsc{Adaptive Difficulty Sampling}}
    \label{alg:adaptive_gfpo}
    \begin{algorithmic}[1]
      \Require group $\mathcal{G}=\{o_1,\dots,o_{G}\}$,  
               t-digest tracker $\mathcal{T}$,
               reward function $R(\cdot)$, prompt $q$,
               scoring $\mathit{metric}(\cdot)$,
                sort $\mathit{order}$ ($\uparrow$/$\downarrow$)

      \State $\mu_R \leftarrow \frac1{G}\sum_{i=1}^{G} R(q, o_i)$ 
      \State $\mathcal{T}.\textsc{Update}(\mu_R)$ 
      \If{$\mathcal{T}.\textsc{Ready}()$}
          \State $(q_{25},q_{50},q_{75})\leftarrow \mathcal{T}.\textsc{Percentile}([25, 50, 75])$ 
                        % \mathcal{T}.\textsc{Percentile}(50), 
                        % \mathcal{T}.\textsc{Percentile}(75)\bigr)$
          \State \textbf{if} $\mu_R < q_{25}$ \textbf{then} $k \leftarrow k_{\text{very-hard}}$
          \State \textbf{elif} $\mu_R < q_{50}$ \textbf{then} $k \leftarrow k_{\text{hard}}$
          \State \textbf{elif} $\mu_R < q_{75}$ \textbf{then} $k \leftarrow k_{\text{med}}$
          \State \textbf{else} $k \leftarrow k_{\text{easy}}$
      \Else
          \State $k \leftarrow k_{\text{very-hard}}$ 
      \EndIf

      \State $(\mathcal{S},m)\gets\textsc{RejectionSample}\bigl(\mathcal{G},k,\mathit{metric},\mathit{order})$
      \State \Return $\mathcal{S},\;m$
    \end{algorithmic}
  \end{algorithm}
  \end{minipage}
  \vspace{-1\baselineskip}
\end{wrapfigure}

%% file: sections/setup.tex
\section{Setup}
\textbf{Model.} We demonstrate the effectiveness of GFPO via Phi-4-reasoning~\citep{abdin2025phi4reasoningtechnicalreport} as the base model. This model was derived by extensive supervised fine-tuning of the 14 billion parameter Phi-4~\citep{abdin2024phi4technicalreport} model on synthetically generated o3-mini reasoning traces primarily on STEM domains, but has not been tuned with any RL. In our results and analysis, we refer to Phi-4-reasoning as the SFT baseline.

\textbf{Baseline.} We compare our GFPO-tuned models with the GRPO trained baseline Phi-4-reasoning-plus~\citep{abdin2025phi4reasoningtechnicalreport}. Note that we use GRPO with the DAPO token-level loss aggregation and slightly modify the clipped surrogate objective to improve training stability, as detailed in Section \ref{sec:gfpo}. We replicate the training setup of Phi-4-reasoning-plus, as detailed below. In our results and analysis, we refer to Phi-4-reasoning-plus as the GRPO baseline.

\textbf{Dataset.} Our RL training focuses on improving mathematical reasoning. The training dataset contains solely of $72k$ math problems selected from a larger training corpus~\citep{abdin2025phi4reasoningtechnicalreport}. Notably, we constrain RL training to 100 steps with batch size of 64 so the model only sees $6.4k$ of these problems during training---identical to the set used to train Phi-4-reasoning-plus.

\textbf{Reward Function.} We use the reward function used for training Phi-4-reasoning-plus~\citep{abdin2025phi4reasoningtechnicalreport}. The reward is a weighted sum of a 0/1 accuracy reward---scaled to be ``length-aware''---and an n-gram repetition penalty. The binary accuracy reward $R_{\text{acc}}$ is computed by extracting the final answer from the response and verifying its equivalence with the ground truth answer and deferring to LLM verifiers if simple answer extraction fails. This reward is then scaled to a float between $-1.0$ and $1.0$ based on the response length, penalizing long responses for correct answers. Formatting violations are penalized by receiving the lowest reward. The final reward function $R$ is a weighted combination of this length-aware accuracy reward and a repetition penalty based on repetition frequency of 5-grams. 
\begin{align}
R =w_{\text{acc}}\textsc{LengthScale}(R_{\text{acc}}) + w_{\text{rep}} R_{\text{rep}},
\end{align}
where $R_{\text{acc}} \in \{0, 1\}$ and $R \in [-1, 1]$.
See Section 4.1 of \cite{abdin2025phi4reasoningtechnicalreport} for more details. Notably, the length penalty in the reward proves insufficient to curb the response length inflation caused by GRPO, motivating our proposal of GFPO.

\textbf{Training Configuration.} We use the \texttt{verl}~\citep{sheng2024hybridflow} framework for GFPO training using the specified reward function. Matching the training setup of Phi-4-reasoning-plus, we GFPO-tune Phi-4-reasoning on 32 H100s with a global batch size of 64 across GPUs for 100 steps, with the Adam optimizer with learning rate of 1e-7\footnote{Phi-4-reasoning-plus is trained with a slightly lower learning rate of 5e-8, but we replicate this run with learning rate of 1e-7 and use the same for GFPO.}, cosine warmup for the first 10 steps, and sampling temperature $T=1.0$. We apply KL regularization with $\beta=0.001$ and an entropy coefficient of $\gamma=0.001$. Our models are trained with $32k$ maximum context length, with $1k$ tokens reserved for the prompt.

\textbf{Group Size.} Phi-4-reasoning-plus is trained with a GRPO group size of $G = 8$. GFPO increases $G$ to increase exposure to desirable responses, trading off more training-time compute for less inference-time compute resulting from shorter responses. We experiment with $G \in {8, 16, 24}$ for GFPO, but the GFPO retained group size $k = |\mathcal{S}| \leq 8$ for all experiments to match the number of responses for which the model receives policy gradient signals for a fair comparison with GRPO.

\textbf{Evaluation.}
We evaluate our checkpoints on AIME 25~\citep{AIME25} and 24~\citep{AIME8324} with 32 samples per prompt, GPQA~\citep{rein2024gpqa} with 5 samples, Omni-MATH~\citep{gao2024omni} with 1 sample, and LiveCodeBench (8/24-1/25)~\citep{jain2024livecodebench} with 3 samples. We sample responses at temperature $T=0.8$ with maximum length of $32k$ with $1k$ tokens reserved for the prompt. For AIME 25, 24, GPQA, and Omni-MATH we first use regex based answer extraction and then use GPT-4o for LLM based extraction if regex extraction fails. While there was no coding data in our RL training set, we evaluate on LiveCodeBench to measure how optimizing for shorter responses with GFPO impacts responses lengths and accuracy out-of-distribution.

We report the average pass@1 accuracy, raw response lengths ($L$), and excess length reduction ($ELR$) for all models and datasets (Tables \ref{tab:table_1}, \ref{tab:table_2}). We define \textit{excess length reduction}, the extent by which GFPO reduces response length inflation caused by GRPO over the SFT model, as follows:
\begin{align}
ELR = \frac{L_{GRPO} - L_{GFPO}}{L_{GRPO} - L_{SFT}}
\label{eq:elr}
\end{align}
\input{tables/acc_len_red_table1}To understand whether GFPO accuracies match those of GRPO, we assess whether the gap in pass@1 accuracies between GRPO and the GFPO variants is meaningful using the Wilcoxon signed-rank test~\citep{wilcoxon1992individual}. This non-parametric, paired test compares the per-question differences in pass@1 accuracy without assuming a normal distribution.

%% file: tables/acc_len_red_table1.tex
\begin{table}[t]
\centering
\begin{tabular}{@{}lllcllcllc@{}}
\toprule
 & \multicolumn{3}{c}{\textbf{AIME 25}} & \multicolumn{3}{c}{\textbf{AIME 24}} & \multicolumn{3}{c}{\textbf{GPQA}} \\ \midrule
 & \textbf{Acc} & \textbf{\begin{tabular}[c]{@{}l@{}}Avg \\ Len\end{tabular}} & \multicolumn{1}{c|}{\textbf{\begin{tabular}[c]{@{}c@{}}\% Len\\ Inf ($\downarrow$)\end{tabular}}} & \textbf{Acc} & \textbf{\begin{tabular}[c]{@{}l@{}}Avg\\ Len\end{tabular}} & \multicolumn{1}{c|}{\textbf{\begin{tabular}[c]{@{}c@{}}\% Len \\ Inf ($\downarrow$)\end{tabular}}} & \textbf{Acc} & \textbf{\begin{tabular}[c]{@{}l@{}}Avg\\ Len\end{tabular}} & \textbf{\begin{tabular}[c]{@{}c@{}}\% Len \\ Inf ($\downarrow$)\end{tabular}} \\ \midrule
\rowcolor[HTML]{CFCFCF} 
SFT & {\color[HTML]{000000} 64.2} & {\color[HTML]{000000} 10.9k} & \multicolumn{1}{c|}{\cellcolor[HTML]{CFCFCF}{\color[HTML]{000000} N/A}} & {\color[HTML]{000000} 72.2} & {\color[HTML]{000000} 10.1k} & \multicolumn{1}{c|}{\cellcolor[HTML]{CFCFCF}{\color[HTML]{000000} N/A}} & {\color[HTML]{000000} 67.0} & {\color[HTML]{000000} 6.6k} & {\color[HTML]{000000} N/A} \\
\rowcolor[HTML]{CFCFCF} 
GRPO & {\color[HTML]{000000} 72.4} & {\color[HTML]{000000} 14.8k} & \multicolumn{1}{c|}{\cellcolor[HTML]{CFCFCF}{\color[HTML]{000000} 0.0}} & {\color[HTML]{000000} 77.7} & {\color[HTML]{000000} 13.3k} & \multicolumn{1}{c|}{\cellcolor[HTML]{CFCFCF}{\color[HTML]{000000} 0.0}} & {\color[HTML]{000000} 67.5} & {\color[HTML]{000000} 10.7k} & {\color[HTML]{000000} 0.0} \\ \midrule
6 of 8 & {\color[HTML]{000000} 69.2} & {\color[HTML]{000000} 14.7k} & \multicolumn{1}{c|}{{\color[HTML]{000000} 1.8}} & {\color[HTML]{000000} 79.6} & {\color[HTML]{000000} 13k} & \multicolumn{1}{c|}{{\color[HTML]{000000} 9.5}} & {\color[HTML]{000000} 70.2} & {\color[HTML]{000000} 10.2k} & {\color[HTML]{000000} 11.5} \\ \midrule
8 of 16 & {\color[HTML]{000000} \textbf{70.2}} & {\color[HTML]{000000} 13.9k} & \multicolumn{1}{c|}{{\color[HTML]{000000} 23.8}} & {\color[HTML]{000000} \textbf{77.9}} & {\color[HTML]{000000} 12.3k} & \multicolumn{1}{c|}{{\color[HTML]{000000} 33.0}} & {\color[HTML]{000000} \textbf{70.0}} & {\color[HTML]{000000} 9.7k} & {\color[HTML]{000000} 23.7} \\
6 of 16 & {\color[HTML]{000000} 70.1} & {\color[HTML]{000000} 13.8k} & \multicolumn{1}{c|}{{\color[HTML]{000000} 25.6}} & {\color[HTML]{000000} 76.9} & {\color[HTML]{000000} 12.2k} & \multicolumn{1}{c|}{{\color[HTML]{000000} 35.6}} & {\color[HTML]{000000} 68.3} & {\color[HTML]{000000} 9.1k} & {\color[HTML]{000000} 38.8} \\
4 of 16 & {\color[HTML]{000000} 69.7} & {\color[HTML]{000000} \textbf{13.3k}} & \multicolumn{1}{c|}{{\color[HTML]{000000} \textbf{38.0}}} & {\color[HTML]{000000} 76.6} & {\color[HTML]{000000} \textbf{11.8k}} & \multicolumn{1}{c|}{{\color[HTML]{000000} \textbf{46.8}}} & {\color[HTML]{000000} 68.6} & {\color[HTML]{000000} \textbf{8.8k}} & {\color[HTML]{000000} \textbf{45.7}} \\ \midrule
8 of 24 & {\color[HTML]{000000} \textbf{70.4}} & {\color[HTML]{000000} \textbf{12.6k}} & \multicolumn{1}{c|}{{\color[HTML]{000000} \textbf{54.4}}} & {\color[HTML]{000000} 75.1} & {\color[HTML]{000000} 11.6k} & \multicolumn{1}{c|}{{\color[HTML]{000000} 52.7}} & {\color[HTML]{000000} 68.9} & {\color[HTML]{000000} 8.6k} & {\color[HTML]{000000} 52.2} \\
6 of 24 & {\color[HTML]{000000} 68.5} & {\color[HTML]{000000} 13.1k} & \multicolumn{1}{c|}{{\color[HTML]{000000} 41.0}} & {\color[HTML]{000000} 75.6} & {\color[HTML]{000000} 11.9k} & \multicolumn{1}{c|}{{\color[HTML]{000000} 44.9}} & {\color[HTML]{000000} \textbf{70.2}} & {\color[HTML]{000000} 8.7k} & {\color[HTML]{000000} 48.6} \\
4 of 24 & {\color[HTML]{000000} 70.3} & {\color[HTML]{000000} 13k} & \multicolumn{1}{c|}{{\color[HTML]{000000} 46.1}} & {\color[HTML]{000000} \textbf{76.5}} & {\color[HTML]{000000} \textbf{11.3k}} & \multicolumn{1}{c|}{{\color[HTML]{000000} \textbf{59.8}}} & {\color[HTML]{000000} 68.1} & {\color[HTML]{000000} \textbf{8.3k}} & {\color[HTML]{000000} \textbf{57.3}} \\ \midrule
\begin{tabular}[c]{@{}l@{}}Token \\ Efficiency\end{tabular} & {\color[HTML]{000000} 69.5} & {\color[HTML]{000000} \textbf{12k}} & \multicolumn{1}{c|}{{\color[HTML]{000000} \textbf{70.9}}} & {\color[HTML]{000000} 76.4} & {\color[HTML]{000000} \textbf{10.6k}} & \multicolumn{1}{c|}{{\color[HTML]{000000} \textbf{84.6}}} & {\color[HTML]{000000} 68.5} & {\color[HTML]{000000} \textbf{7.5k}} & {\color[HTML]{000000} \textbf{79.7}} \\ \midrule
\begin{tabular}[c]{@{}l@{}}Adaptive \\ Difficulty\end{tabular} & {\color[HTML]{000000} \textbf{70.8}} & {\color[HTML]{000000} 12.8k} & \multicolumn{1}{c|}{{\color[HTML]{000000} 50.8}} & {\color[HTML]{000000} \textbf{76.6}} & {\color[HTML]{000000} 11.6k} & \multicolumn{1}{c|}{{\color[HTML]{000000} 52.9}} & {\color[HTML]{000000} \textbf{70.8}} & {\color[HTML]{000000} 9k} & {\color[HTML]{000000} 41.7} \\ \bottomrule
\end{tabular}
\caption{\textbf{Pass@1 Accuracy, Response Lengths, and Length Inflation Reduction on AIME 25, AIME 24, and GPQA.} 
GFPO variants match GRPO accuracy (no statistically significant differences under the Wilcoxon signed-rank test) while reducing length inflation across all benchmarks. 
Across configurations, sampling more responses is key, and lowering the $k/G$ ratio is an effective lever for controlling length. 
Token Efficiency delivers the largest reductions overall, while Adaptive Difficulty surpasses Shortest $k/G$ at equivalent compute.
Pass@1 accuracy is computed over 32 samples for AIME 25/24 and 5 samples for GPQA. We \textbf{highlight} the best accuracy and length within $G=16$ and $G=24$ and between Token Efficiency and Adaptive Difficulty. See Table~\ref{tab:table_2} for results on Omni-MATH, LiveCodeBench, and average performance across all benchmarks.}
\label{tab:table_1}
\end{table}

%% file: sections/results.tex
\section{Results}
We evaluate three GFPO variants:
\begin{itemize}[leftmargin=2em]
\item \textbf{Shortest $k/G$}: retains the $k$ shortest responses from $\mathcal{G}$, with both $k$ and group size $G$ varied to study their effect on length reduction.
\item \textbf{Token Efficiency}: retains the $k$ most reward-per-token efficient responses from $\mathcal{G}$, using $k=8$, $G=16$ (matching the baseline Shortest $k/G$ setting).
\item \textbf{Adaptive Difficulty}: retains the $k$ shortest responses from $\mathcal{G}$, with $k$ chosen dynamically with real-time difficulty estimates (4, 6, 8, and 8 for easy→very hard) and $G=16$.
\end{itemize}
We measure pass@1 accuracy and Excess Length Reduction (Equation~\ref{eq:elr}). The Wilcoxon signed-rank test shows no statistically significant accuracy differences between GFPO variants and GRPO across tasks, indicating GFPO preserves accuracy while reducing length.

\mySubsection{\textbf{Think Less Without Sampling More?}}
A natural question is whether rejection sampling alone—without increasing the total number of sampled responses—can significantly reduce response lengths. To study this, we experiment with Shortest 6/8 GFPO---subsampling the $k=6$ shortest responses from a group size of $G=8$. We find that Shortest 6/8 GFPO achieves comparable accuracy to GRPO on AIME 25, AIME 24, GPQA, and Omni-MATH. However, the resulting reductions in excess length are modest: Shortest 6/8 GFPO only cuts length inflation by 1.8\%, 9.5\%, and 11.5\% on AIME 25, AIME 24, and GPQA respectively and even leads to a minor length increase on Omni-MATH (+5.5\%) (Tables~\ref{tab:table_1},~\ref{tab:table_2}).\input{tables/acc_len_red_table2}This suggests that while subsampling from a small group can yield modest reductions in response length, more substantial reductions may require increasing G to draw shorter chains to train on. This motivates our subsequent experiments exploring if \textit{sampling more} can enable models to \textit{think less}.

\mySubsection{\textbf{GFPO Enables Efficient Reasoning.}}\label{res:grpo_efficient}
Motivated by the observation that sampling more responses may substantially reduce chain length, we investigate the effectiveness of the Shortest 8/16 GFPO variant. In this method, we increase our group sample size to $G=16$ and retain only the shortest 8 responses—effectively training the model on the shortest 50\% of sampled chains, with the rejected samples receiving zero advantage.
\begin{findings}
\textbf{``Thinking Less'' Requires Sampling More: }Reducing retained responses without increasing group size (Shortest 6/8 GFPO) does not reduce response length. 
\end{findings}
Applying Shortest 8/16 GFPO, we observe significant reductions in length inflation across multiple benchmarks---23.8\% reduction in excess length on AIME 25, 33\% reduction on AIME 24, 23.7\% reduction on GPQA, and 31.5\% reduction on Omni-MATH, all without any statistically significant drops in accuracy (Tables~\ref{tab:table_1},~\ref{tab:table_2}). Overall, GFPO substantially reduces response lengths while maintaining the strong reasoning performance of GRPO.

\mySubsection{\textbf{Effect of Varying $k$ and $G$ on Length Reductions.}}\label{res:vary_k_G} We next investigate how varying the retained group size ($k$) and the sampled group size ($G$) affects length reduction, motivated by the intuition that either rejecting more responses ($k\downarrow$) or sampling more extensively ($G\uparrow$) could further shorten reasoning chains.
\input{figures/pareto_plot_wrap}
Compared to Shortest 8/16 GFPO, slightly reducing the retained set to Shortest 6/16 GFPO yields moderate additional reductions: 1.8\% more on AIME 25, 2.6\% more on AIME 24, 15.1\% more on GPQA, and 12.2\% more on Omni-MATH. Decreasing $k$ further with Shortest 4/16 GFPO achieves even stronger improvements over Shortest 8/16, providing 14.2\%, 13.8\%, 22\%, and 15.8\% additional excess length reduction on the same benchmarks (Tables~\ref{tab:table_1},~\ref{tab:table_2}). 

We also investigate scaling up the sampled group size $G$ from 16 to 24, while holding $k$ fixed. Moving from Shortest 8/16 to 8/24 yields substantial additional reductions in excess length (30.6\%, 19.7\%, 28.5\%, and 20.4\% more on AIME 25, AIME 24, GPQA, and Omni-MATH respectively). Similarly, moving from Shortest 6/8 to 6/16 \input{figures/len_vs_k_G_wrap}provides large added excess length reductions (23.4\%, 26.2\%, 27.3\%, and 49.2\% more), with smaller additional improvements (15.4\%, 9.3\%, 9.8\%, and 14.5\% more) when scaling further to 6/24. Finally, increasing from Shortest 4/16 to 4/24 achieves additional length reductions of 8.1\%, 13\%, 11.5\%, and 23.7\% respectively on the same datasets (Tables~\ref{tab:table_1},~\ref{tab:table_2}).

These results collectively indicate that the crucial knob for controlling response length is the proportion of retained responses ($k/G$)---decreasing $k/G$ by decreasing $k$ or increasing $G$ enables response length reductions (Figure~\ref{fig:len_vs_k_G}). We confirm this by comparing two configurations—Shortest 4/16 and Shortest 6/24—with identical retention fractions of 25\% but different absolute values of $k$ and $G$. Shortest 6/24 achieves slightly better average length reductions (by 2.9\%) (Table~\ref{tab:table_2}, Figure~\ref{fig:len_vs_k_G}), reflecting that a larger sampled group size ($G$) increases the chance of encountering high-quality short chains. Notably, this suggests that once the sampled group size is sufficiently large ($\uparrow G$ from $8\rightarrow16$), strong length reductions can be efficiently achieved by simply tuning $k$ without needing further sampling.
Ultimately, decreasing the retained fraction too aggressively yields marginal improvements (e.g., 8/24 to 4/24 yields only 4.1\% average additional excess length reduction) (Table~\ref{tab:table_2}). Therefore, to push beyond these limits, we must leverage smarter sampling strategies.

\begin{findings}
\textbf{\% of Retained Responses ($k/G$) Controls Length Pressure:} Lowering $k$ or raising $G$ further reduces length; retaining 25–33\% of responses is observed as optimal, with smaller ratios yielding diminishing gains.
Shortest 4/24 is the best length optimized GFPO variant, yielding the strongest excess length reductions.
\end{findings}
\mySubsection{\textbf{Reinforcing Token Efficiency.}}\label{res:tok_eff} 
Earlier experiments revealed that simply decreasing the proportion of retained responses $k/G$ eventually hits a ceiling: beyond a certain group size it is difficult to yield substantially shorter reasoning chains. To overcome this plateau, we introduce Token Efficiency GFPO—a “smart-sampling” approach that ranks responses by reward-per-token (reward/length). The intuition is straightforward: the policy should prioritize chains that deliver high reward efficiently; longer solutions should be favored only if justified by high enough rewards.

Mechanically, Token Efficiency GFPO retains the top-$k$ responses according to the ratio $R_i/|o_i|$. Short correct chains and some long correct and incorrect chains maximize this ratio. Computing advantages within this filtered set results in short correct chains receiving the strongest positive gradients, long correct chains receiving modest bonuses or mild penalties, and long incorrect chains incurring the steepest penalties. The additional gradient pressure on long incorrect chains trims ``filler'' tokens that shortest-$k$ GFPO cannot directly target, as it provides no gradient signal beyond the length of the longest retained chain. While shortest-$k$ relies on the KL penalty to implicitly nudge late-token probabilities downward, Token Efficiency GFPO supplies explicit negative gradients to actively disincentivize these late, low-value token positions.

We train this method with $k=8$ and $G=16$. Token Efficiency GFPO delivers the largest excess length reductions across all tasks—70.9\% on AIME 25, 84.6\% on AIME 24, 79.7\% on GPQA, and 82.6\% on Omni-MATH—outperforming the shortest-$k$ variants (Tables~\ref{tab:table_1},~\ref{tab:table_2}) smaller or equivalent $G$. These added length reductions come at a slight cost: training curves exhibit higher variance in policy performance (Figure~\ref{fig:aime_training}) and we observe minor non-statistically significant degradations in accuracy (Tables~\ref{tab:table_1},~\ref{tab:table_2}). 
\begin{findings}
\textbf{Token-efficiency ($reward/length$) Optimization Yields Largest Cuts:} Excess length reductions of \textbf{70.9\%} (AIME 25), \textbf{84.6\%} (AIME 24), \textbf{79.7\%} (GPQA), \textbf{82.6\%} (Omni-MATH), \textbf{79.7\%} (LiveCodeBench) while maintaining accuracy. These reductions come with slightly increased variance during training.
\end{findings}
This variance is likely due to noisy gradients on token segments that occur in both long correct and incorrect responses, causing conflicting reward and penalty signals. Nevertheless, Token Efficiency GFPO achieves the strongest token savings without sacrificing accuracy, confirming that reward-per-token is an effective proxy for genuinely concise reasoning.

\mySubsection{\textbf{Adaptive Difficulty GFPO.}}\label{res:adaptive_diff} Beyond intelligent sampling through improved rejection metrics, we introduce Adaptive Difficulty GFPO, a method for strategically determining the retained group size $k$ based on question difficulty—allocating more training resources to harder questions.

In Adaptive Difficulty GFPO (Section~\ref{sec:adaptive_diff_method}), we estimate question difficulty using the average reward of responses per question, efficiently compute problem difficulty quartiles at each training step, and categorize questions into four difficulty buckets: very hard (bottom 25\%), hard (25--50\%), medium (50--75\%), and easy (top 25\%). Based on this categorization, we retain the 8, 8, 6, or 4 shortest responses (of $G=16$ sampled) for questions from hardest to easiest, respectively (Algorithm~\ref{alg:adaptive_gfpo}).

For this configuration of Adaptive Difficulty GFPO, the average $k$ per question is 6.5, so we compare this approach against Shortest 6/16 GFPO which closely matches the number of retained responses $k$ and the group size $G$. While Shortest 6/16 GFPO achieves stronger excess length reductions on Omni-MATH (43.7\% vs. 35.1\%), Adaptive Difficulty GFPO outperforms it on AIME 25 (50.8\% vs. 25.6\%), AIME 24 (52.9\% vs. 35.6\%), and GPQA (41.7\% vs. 38.8\%), despite Shortest 6/16 GFPO applying slightly more aggressive response pruning. Compared to Shortest 4/16 GFPO, which filters even more aggressively, Adaptive Difficulty GFPO still achieves superior excess length reductions on AIME 25 (50.8\% vs. 38\%) and AIME 24 (52.9\% vs. 46.8\%) (Tables~\ref{tab:table_1},~\ref{tab:table_2}).
\begin{findings}
\textbf{Adaptive Difficulty GFPO Beats Shortest-k at Equal Compute}: Adaptively deciding $k$ based on question difficulty yields stronger length reductions on 4/5 benchmarks than Shortest-$k$ at equivalent compute.
\end{findings}
Adaptive Difficulty GFPO also attains the highest accuracy on GPQA (70.8\%) (Table~\ref{tab:table_1}) and on the hardest AIME 25 questions (27\%) (Figure~\ref{fig:acc_vs_diff_quartile}) compared to GRPO and all GFPO variants.  These results underscore the effectiveness of strategically allocating sampling budget based on problem difficulty. Note that Adaptive Difficulty GFPO could be optimized with the token efficiency metric for even stronger outcomes.

\mySubsection{\textbf{Out-of-Distribution Effects of GFPO.}}\label{res:ood} Our RL training recipe is geared towards enhancing mathematical reasoning performance. To investigate potential adverse effects of GFPO's bias toward shorter responses, we assess out-of-distribution generalization on the LiveCodeBench coding benchmark. Note that coding is not a part of our RL training set.

We observe that GRPO leads to significant response length inflation even out-of-distribution, increasing average response length from 10.3\textit{k} tokens (SFT) to 13.9\textit{k} tokens without improving accuracy (56.7\% GRPO vs 57.7\% SFT). While thinking for longer as a result of RL may be warranted to solve harder problems in-distribution, this length inflation is unexpected and undesirable for out-of-distribution tasks, particularly when no accuracy gains accompany the longer outputs.
\begin{findings}
\textbf{GFPO Mitigates OOD Length Inflation}: GRPO increases response length on out-of-distribution tasks without accuracy gains; GFPO curbs this while modestly improving accuracy.
\end{findings}
GFPO effectively mitigates this unintended verbosity. On LiveCodeBench, Token Efficiency GFPO achieves the most substantial reduction (79.7\%) in excess response length. GFPO variants even yield modest accuracy improvements on the coding task: Shortest 8/24 GFPO slightly outperforms both SFT and GRPO in accuracy (59.2\% vs. 57.7\% and 56.7\%, respectively) while simultaneously cutting excess length by 57\%. These results highlight GFPO’s ability to maintain—and even slightly enhance—out-of-distribution generalization while explicitly managing response length increase.

\mySubsection{\textbf{Accuracy-Length Pareto Comparison.}} 
Figure \ref{fig:pareto_plots} provides a holistic view of the accuracy---response length frontier.  
Across four of the five benchmarks, at least one GFPO variant is strictly \textit{Pareto-superior} to GRPO (landing in the green  region), confirming that GFPO can deliver both shorter and more accurate answers, improving both axes simultaneously.  
Even on AIME 25, where GRPO has slightly better accuracy, GFPO variants remain on the Pareto front by offering meaningful length reductions without statistically significant losses in accuracy.  
Aggregating across tasks (bottom-right panel) highlights Shortest 4/24, Adaptive Difficulty, and Shortest 8/16 as the most consistently concise and accurate methods, with Token Efficiency trailing in accuracy by a narrow margin.

%% file: tables/acc_len_red_table2.tex
\begin{table}[t]
\centering
\begin{tabular}{@{}lccccccccc@{}}
\toprule
 & \multicolumn{3}{c}{\textbf{Omni-MATH}} & \multicolumn{3}{c}{\textbf{LiveCodeBench}} & \multicolumn{3}{c}{\textbf{Average}} \\ \midrule
 & \textbf{Acc} & \textbf{\begin{tabular}[c]{@{}c@{}}Avg \\ Len\end{tabular}} & \multicolumn{1}{c|}{\textbf{\begin{tabular}[c]{@{}c@{}}\% Len \\ Inf ($\downarrow$)\end{tabular}}} & \textbf{Acc} & \textbf{\begin{tabular}[c]{@{}c@{}}Avg \\ Len\end{tabular}} & \multicolumn{1}{c|}{\textbf{\begin{tabular}[c]{@{}c@{}}\% Len \\ Inf ($\downarrow$)\end{tabular}}} & \textbf{Acc} & \textbf{\begin{tabular}[c]{@{}c@{}}Avg \\ Len\end{tabular}} & \textbf{\begin{tabular}[c]{@{}c@{}}\% Len\\ Inf ($\downarrow$)\end{tabular}} \\ \midrule
\rowcolor[HTML]{CFCFCF} SFT & 84.7 & 9.6k & \multicolumn{1}{c|}{N/A} & 57.7 & 10.3k & \multicolumn{1}{c|}{N/A} & 69.2 & 9.5k & N/A \\
\rowcolor[HTML]{CFCFCF} GRPO & 86.0 & 12.7k & \multicolumn{1}{c|}{0.0} & 56.7 & 13.9k & \multicolumn{1}{c|}{0.0} & 72.1 & 13k & 0.0 \\ \midrule
6 of 8 & 88.3 & 12.9k & \multicolumn{1}{c|}{-5.5} & 56.4 & 13.6k & \multicolumn{1}{c|}{7.0} & 72.7 & 12.9k & 4.8 \\ \midrule
8 of 16 & \textbf{89.3} & 11.8k & \multicolumn{1}{c|}{31.5} & \textbf{59.8} & 12.6k & \multicolumn{1}{c|}{36.5} & \textbf{73.4} & 12k & 29.7 \\
6 of 16 & 87.8 & 11.4k & \multicolumn{1}{c|}{43.7} & 58.3 & \textbf{12.6k} & \multicolumn{1}{c|}{\textbf{37.2}} & 72.3 & 11.8k & 36.2 \\
4 of 16 & 88.0 & \textbf{11.3k} & \multicolumn{1}{c|}{\textbf{47.3}} & 57.2 & 12.3k & \multicolumn{1}{c|}{43.2} & 72.0 & \textbf{11.5k} & \textbf{44.2} \\ \midrule
8 of 24 & 87.5 & 11.1k & \multicolumn{1}{c|}{51.9} & 56.5 & \textbf{11.8k} & \multicolumn{1}{c|}{\textbf{59.4}} & 71.7 & 11.1k & 54.1 \\
6 of 24 & \textbf{88.1} & 10.9k & \multicolumn{1}{c|}{58.2} & 58.7 & 12.4k & \multicolumn{1}{c|}{42.7} & 72.2 & 11.4k & 47.1 \\
4 of 24 & 87.6 & \textbf{10.5k} & \multicolumn{1}{c|}{\textbf{71.0}} & \textbf{59.2} & 11.8k & \multicolumn{1}{c|}{57.0} & \textbf{72.3} & \textbf{11k} & \textbf{58.2} \\ \midrule
\begin{tabular}[c]{@{}l@{}}Token\\ Efficiency\end{tabular} & 87.4 & \textbf{10.1k} & \multicolumn{1}{c|}{\textbf{82.6}} & 57.0 & \textbf{11k} & \multicolumn{1}{c|}{\textbf{79.7}} & 71.7 & \textbf{10.2k} & \textbf{79.5} \\ \midrule
\begin{tabular}[c]{@{}l@{}}Adaptive\\ Difficulty\end{tabular} & \textbf{88.9} & 11.6k & \multicolumn{1}{c|}{35.1} & \textbf{57.2} & 12.1k & \multicolumn{1}{c|}{49.4} & \textbf{72.9} & 11.4k & 46.0 \\ \bottomrule
\end{tabular}
\caption{\textbf{Pass@1 Accuracy, Response Lengths, and Length Inflation Reduction on Omni-MATH, LiveCodeBench, and Average Across All Benchmarks.} 
GFPO variants cut length inflation and slightly boost accuracy over GRPO on Omni-MATH. 
On LiveCodeBench (out-of-distribution, coding), GRPO exhibits substantial length inflation without accuracy gains, whereas GFPO variants reduce length and in some cases improve accuracy (e.g., 8/16, 4/24). 
On average across all benchmarks (AIME 25, AIME 24, GPQA, Omni-MATH, LiveCodeBench), Token Efficiency GFPO achieves the largest excess length reduction (79.5\%) while maintaining GRPO-level accuracy. 
Pass@1 accuracy is computed over 1 sample for Omni-MATH and 3 samples for LiveCodeBench.}
\label{tab:table_2}
\end{table}

%% file: figures/pareto_plot_wrap.tex
\begin{figure}[t]
    \centering
    \includegraphics[width=0.9\textwidth]{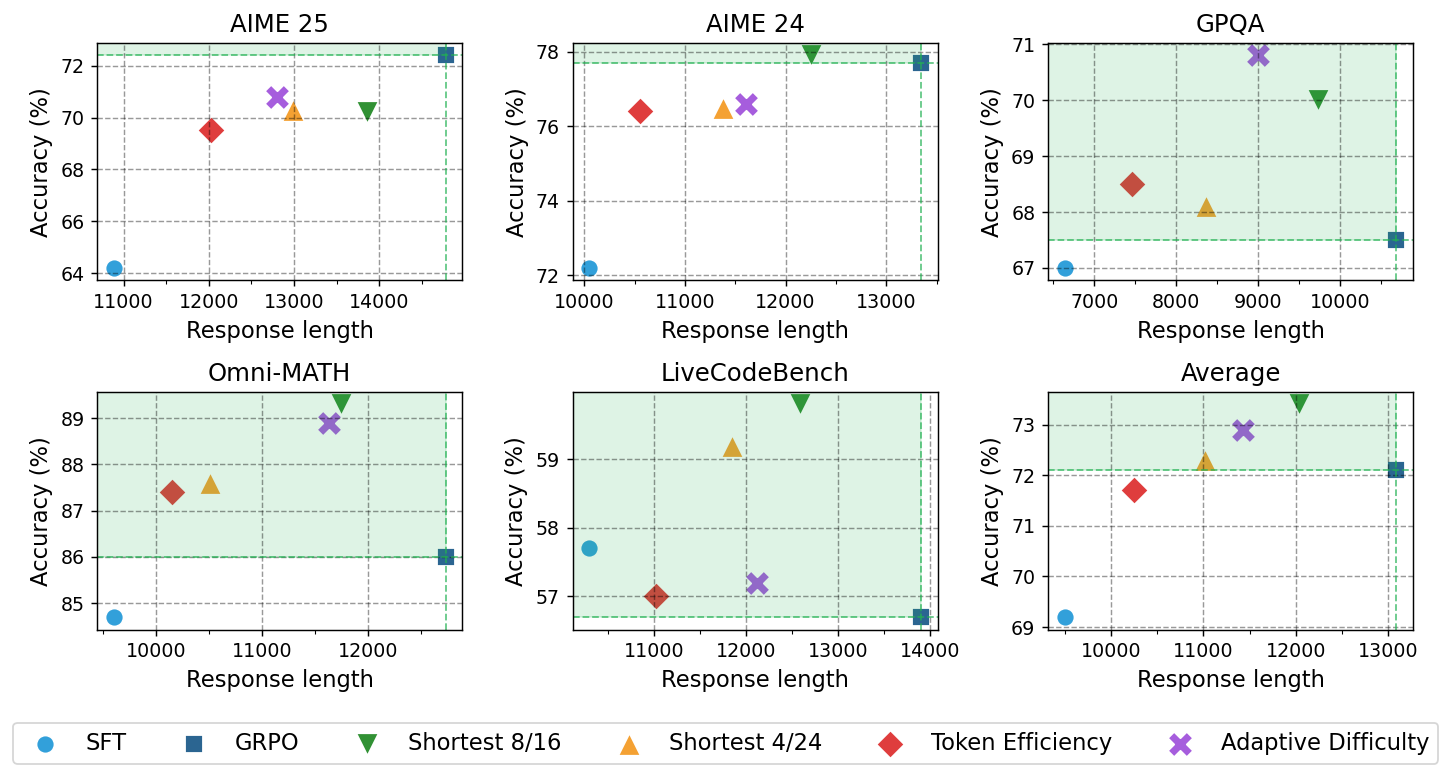}
    \caption{\textbf{Pareto Trade-off Between Accuracy and Response Length. } For all benchmarks except AIME 25, at least one GFPO variant strictly dominates GRPO—achieving both higher accuracy and shorter responses (green region above and to the left of GRPO). For AIME 25, GRPO attains the highest accuracy, but several GFPO variants, while taking non-significant accuracy dips, remain Pareto-optimal because their responses are shorter, and no other method is simultaneously more accurate and more concise. On average, Shortest 4/24, Adaptive Difficulty, and Shortest 8/16 are strictly Pareto-superior to GRPO with Token Efficiency close behind.}
    \label{fig:pareto_plots}
    \vspace{-1\baselineskip}
\end{figure}

%% file: figures/len_vs_k_G_wrap.tex
\begin{wrapfigure}{r}{0.4\textwidth}
  \vspace{-0.8\baselineskip}
  \centering
  \includegraphics[width=0.4\textwidth]{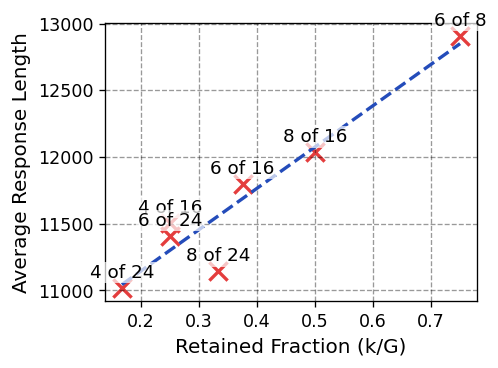}
  \caption{\textbf{Average Response Length vs k/G.} $\downarrow k/G \rightarrow \downarrow L_{avg}$ but beyond a point leads to diminishing returns.}
  \label{fig:len_vs_k_G}
  \vspace{-1.5\baselineskip}
\end{wrapfigure}

%% file: sections/analysis.tex
\section{Analysis}\label{sec:analysis}
\input{figures/len_red_problem_diff_wrap}
We analyze GFPO’s behavior on the AIME 2025 dataset by measuring question difficulty as $1 -$ SFT accuracy—capturing how challenging each problem is for the base SFT model prior to RL. Questions are partitioned into difficulty quartiles (easy, medium, hard, very hard) to study how GFPO affects response length and accuracy across difficulty levels. We then examine the accuracy of long responses under fixed difficulty and how GFPO reshapes the joint length–accuracy distribution. Finally, we investigate what parts of responses GFPO is trimming and share qualitative examples comparing GFPO with GRPO in Appendix~\ref{appendix:qual_ex}.

\mySubsection{\textbf{Length Reductions on Easy vs Hard Problems.}} We analyze how GFPO's length reductions vary with question difficulty on AIME 2025. As expected, response lengths significantly increase with difficulty—from roughly $4k$ tokens on easy questions to over $20k$ tokens on very hard ones (Figure~\ref{fig:len_over_diff}). GFPO effectively reduces lengths across all difficulty levels (Figure~\ref{fig:len_red_over_diff}).

Token Efficiency GFPO achieves the strongest overall reductions, particularly on easy questions (121.6\% excess length reduction compared to GRPO) (Figure~\ref{fig:len_red_over_diff}), surpassing even the SFT model’s brevity while improving accuracy—demonstrating that length and accuracy can be optimized simultaneously. However, its reductions decrease on harder questions (79.1\% on medium, 63.5\% on hard, 56.5\% on very hard) because the token efficiency metric allows longer responses when justified by higher rewards, common for difficult problems requiring additional reasoning.

Adaptive Difficulty GFPO exhibits increasing excess length reductions with difficulty (37.7\% on easy vs. 60.3\% on very hard), effectively trimming the "long tail" of lengthy responses (Figure~\ref{fig:len_red_over_diff}).
Although both Adaptive Difficulty and Shortest 8/16 GFPO retain 8 shortest responses for hard problems, Adaptive Difficulty achieves stronger length reductions. This succinctness likely stems from brevity gradients from easier problems, teaching the policy to avoid unnecessary tokens even on challenging tasks.
\input{figures/analysis_second_figs_wrap}

\begin{findings}
\begin{itemize}[leftmargin=2em]
\item \textbf{GFPO shortens responses across all difficulty levels.}
\item Token Efficiency GFPO delivers the largest reductions on easy, medium, and hard questions—on easy questions \textbf{producing responses even shorter than the SFT model while matching GRPO’s accuracy}.
\item Shortest 8/24 GFPO achieves the \textbf{greatest reductions on the hardest questions} due to its stronger filtering.
\end{itemize}
\end{findings}
As expected, Shortest 8/24 GFPO consistently achieves stronger reductions than Shortest 8/16 across all difficulty levels. Notably, Shortest 8/24 yields the largest reductions on very hard questions, outperforming Token Efficiency GFPO—which preserves high-reward long responses—and Adaptive Difficulty GFPO—which retains a larger fraction of responses for hard problems during training (Figure~\ref{fig:len_red_over_diff}).

\mySubsection{\textbf{Accuracy on Easy vs Hard Problems.}} Next, we examine GFPO's accuracy across difficulty levels on AIME 2025 (Figure~\ref{fig:acc_vs_diff_quartile}). All methods achieve near-perfect accuracy (98–99\%) on easy problems. SFT accuracy sharply declines as difficulty increases, whereas RL fine-tuning (GRPO and GFPO) consistently improves performance over SFT. 

Token Efficiency GFPO achieves large length reductions (Figure~\ref{fig:len_red_over_diff}), though with small, statistically insignificant accuracy drops compared to GRPO (Figure~\ref{fig:acc_vs_diff_quartile}).
Adaptive Difficulty GFPO matches or exceeds GRPO’s accuracy on easy, medium, and very hard questions notably improving accuracy on medium problems (90.2\% vs. 88.4\%) while reducing excess length by 47\%. On very hard questions—where other GFPO variants slightly lose accuracy—Adaptive Difficulty matches GRPO (27\% vs. 26.6\%) by adaptively allocating more compute and exploration to challenging problems while simultaneously reducing excess length by 60\% (Figure~\ref{fig:len_red_over_diff}).
\begin{findings}
\begin{itemize}[leftmargin=2em]
\item \textbf{Adaptive Difficulty GFPO surpasses GRPO accuracy on medium and very hard problems} while reducing excess length by 47\%-60\%.
\item \textbf{Larger group sizes improve accuracy on hard problems}: Adaptive Difficulty ($k=8$, $G=16$) drops slightly on hard problems, but Shortest 8/24 matches GRPO accuracy by sampling more to find concise correct answers.
\end{itemize}
\end{findings}

However, Adaptive Difficulty experiences a modest accuracy drop on hard problems. Easy-to-medium problems enable confident filtering of long responses due to all responses having consistently high rewards. Very hard problems rarely yield correct answers even with long responses, enabling aggressive filtering without sacrificing accuracy. \input{figures/relative_len_lineplot_wrap}Hard problems occupy a middle ground, where discarding beneficial longer responses slightly reduces accuracy. Increasing group size (e.g., from 16 to 24 responses per question) can address this limitation.

Shortest 8/24 GFPO illustrates this approach: sampling more responses ensures retaining high-quality concise chains, fully matching GRPO’s accuracy on hard problems (73.4\%) (Figure~\ref{fig:acc_vs_diff_quartile}). Overall, GFPO robustly preserves accuracy across difficulty levels, with Adaptive Difficulty strategically allocating sampling resources to effectively balance length and correctness.

\mySubsection{\textbf{Accuracy of Long Responses under GFPO.}}
Longer responses from reasoning models often show lower accuracy, but this trend is confounded by question difficulty—harder problems naturally elicit longer outputs so accuracy may drop due to question difficulty instead of response verbosity. To disentangle these effects, we hold difficulty constant and analyze how response length alone affects model performance on AIME 2025. We partition each model’s responses for hard and very hard problems into length quartiles (Figure~\ref{fig:acc_rel_len_hard}), using the SFT model’s per-question accuracy as a proxy for difficulty, and compute response accuracy within each quartile  (Figure~\ref{fig:acc_rel_len_hard}).

Accuracy declines consistently with increasing length for both difficulty levels–confirming that longer responses tend to be less accurate, even when problem difficulty is held constant. On hard problems, most models peak at mid-low or mid-high lengths (Figure~\ref{fig:acc_rel_len_hard}) (12k–16k tokens) (Table~\ref{tab:app_len}), suggesting a sweet spot: long enough for reasoning, but short enough to avoid over-thinking.
\begin{findings}
\textbf{Longer responses are less accurate even at fixed difficulty}: Across hard problems, the \textbf{sweet spot for reasoning emerges around 12$k$–16$k$ tokens}.
\end{findings}
Beyond this range, accuracy drops consistently. GFPO variants outperform GRPO in the longest bin in both difficulty levels (66.7\% vs 52.1\% on Hard, 20.3\% vs 17.2\% on Very Hard) (Table~\ref{tab:app_acc}), as their longest responses are less verbose ($20.8k$ vs. $23.8k$ on Hard; $26.9k$ vs. $27.5k$ on Very Hard) (Table~\ref{tab:app_len}) and more accurate than those from GRPO.
 \begin{findings}
\textbf{GFPO outperforms GRPO accuracy in the lengthiest response quartiles.} 
\end{findings}
\input{figures/len_kde_wrap} For very hard problems, accuracy falls off more sharply with length. Only Adaptive Difficulty and Token Efficiency improve from short to mid-low bins; all others decline steadily. Token Efficiency and Shortest 8/24 show the steepest drops in longer bins, likely due to aggressive filtering reducing exposure to long chains in training. Adaptive Difficulty is the most robust, maintaining stable accuracy across bins and avoiding sharp drops in very hard cases. In contrast, SFT—while less prone to degradation—rarely solves hard problems, yielding a flat but low accuracy profile (Figure~\ref{fig:acc_rel_len_hard}).

We complement this with an absolute-length analysis across models on AIME 25 (Figure~\ref{fig:abs_len_heatmap})--holding the response lengths fixed between models and evaluating the accuracy, fraction of responses, and difficulty of the unique prompts corresponding to the responses per fixed length bin. GFPO shifts substantial mass away from the long tail ($\geq$20k tokens), reducing it from 32\% (GRPO) to ~22–23\% and boosting the share of $<$15k responses.
\begin{findings}
\textbf{GFPO cuts extreme verbosity}: dropping the fraction of $\geq20k$-token responses from 32\% to $22\%$, \textbf{while solving harder problems at shorter lengths} (questions answered in $\leq5k$ tokens are 9× harder in GFPO than GRPO).
\end{findings} 
These shorter GFPO responses often solve harder problems: in the $\leq$5k bin, prompt difficulty is ~9× higher than GRPO’s (16–18\% vs. 2\% hardness) with only minor accuracy dips (e.g., 100\% → 97\%). 
Slightly lower accuracy in GFPO’s longest bins reflects that many of these prompts are already solved at shorter lengths; remaining long outputs are rare, out-of-distribution cases corresponding to the hardest questions.
Together, the relative- and absolute-length analyses show verbosity—not difficulty—is the main driver of GRPO’s long-chain errors. GFPO, by contrast, solves difficult problems more succinctly with competitive or better accuracy.

\input{figures/acc_kde_wrap}\mySubsection{\textbf{Distribution-Level Effects of GFPO.}} To visualize how GFPO reshapes the accuracy–length landscape, we plot kernel density estimates for response lengths (Figure~\ref{fig:len_kde}) and pass@1 accuracies over 32 independent runs (Figure~\ref{fig:acc_kde}) on AIME 25. 

We find that all GFPO  variants compress the long tail, shifting mass toward shorter responses (Figure~\ref{fig:len_kde}).
The distribution of pass@1 accuracies on AIME 25 allows a visual comparison of both typical performance and run-to-run variability for each method (Figure~\ref{fig:acc_kde}).
The SFT model’s per-repeat accuracy distribution is left-skewed, indicating lower average performance across runs, while GRPO’s distribution is shifted right, reflecting higher typical accuracies.

GFPO variants fall between these extremes. Token Efficiency GFPO shows the narrowest spread, suggesting the most consistent accuracy from repeat to repeat. Shortest 8/24  closely matches GRPO’s central mass, while Adaptive Difficulty exhibits a slightly bimodal profile—indicating that some repeats achieve GRPO-level accuracy while others drop slightly lower.

We quantify the accuracy–length trade-off on AIME 25 by measuring the share of prompts with accuracy $\geq70\%$ and responses with length $\geq15k$ tokens. GRPO produces the highest proportion of long responses (46.8\%), compared to 28\% for SFT. 
GFPO variants reduce this to 42.1\% (Shortest 8/16), 35\% (Shortest 8/24), 37.8\% (Adaptive Difficulty), and 32.7\% (Token Efficiency). GRPO has 70\% of prompts with accuracy $\geq70\%$, while GFPO variants follow behind, with modest reductions—66.7\% (Shortest 8/16), 60\% (Shortest 8/24), 63.3\% (Adaptive Difficulty), and 60\% (Token Efficiency). 
\input{figures/token_analysis} Note, however, that the differences in average accuracy across prompts between GRPO and GFPO are not statistically significant.

\begin{findings}
\textbf{Which GFPO variant is best?}
\begin{itemize}[leftmargin=2em]
\item \textbf{Token Efficiency}: strongest length reductions, small drops in accuracy
\item \textbf{Adaptive Difficulty}: best accuracy on hardest problems with solid length cuts
\item \textbf{Shortest 8/24}: very effective at managing accuracy-length trade-off 
\end{itemize}
\end{findings}
Overall, both Shortest 8/24 GFPO and Adaptive Difficulty strike a strong balance—substantially shortening responses while preserving high accuracy on a large share of questions. Further gains may be possible by tuning the $k/G$ ratio for Token Efficiency GFPO and Adaptive Difficulty GFPO, or by training Adaptive Difficulty with the token efficiency metric.

\mySubsection{\textbf{What is GFPO trimming?}}
To investigate the source of GFPO’s length reductions, we annotate reasoning traces from five models—SFT, GRPO, Shortest 8/24 GFPO, Token Efficiency GFPO, and Adaptive Difficulty GFPO—on AIME 25 and GPQA using GPT-4o. We first segment each trace into smaller chunks and label every sentence according to its functional role in the reasoning chain:\\
i) \emph{Problem}: text relating to problem formulation and understanding,\\
ii) \emph{Solution}: sentences proposing or developing candidate solutions, \\
iii) \emph{Verification}: steps where the model re-checks or validates intermediate results, and \\
iv) \emph{Final}: the concluding statements that deliver the answer.

A qualitative comparison of GRPO vs GFPO using our annotations is shown below. See Appendix \ref{appendix:qual_ex} for more examples on AIME 25 and GPQA.
\input{figures/qual_example_main_body}
We report the average token counts per section for each model across AIME 25 and GPQA (Figure~\ref{fig:token_analysis_aime_gpqa}). On both benchmarks, GRPO inflates the mid-trace reasoning (Solution and Verification) substantially compared to the SFT baseline. For example, on AIME 25 the Solution segment grows from $6.5k$ tokens (SFT) to $8.3k$ tokens, and Verification from $1.9k$ to $3.1k$ tokens. 

In contrast, the three GFPO strategies effectively compress those phases. Shortest 8/24 GFPO reduces the AIME 25 Solution segment from $8.3k$ to $6.6k$ tokens (94.4\% reduction in excess length). This indicates that GFPO many tokens proposing different candidate solution which may be incorrect or unnecessary digressions, while GFPO hones in on promising candidates and cuts the noise.  Shortest 8/24 also reduces the Verification tokens on AIME 25 from $3.1k$ to $2.3k$ (66.7\% reduction in excess length), cutting a large portion of the repetitive, circular checks characteristic of GRPO.

Token Efficiency GFPO achieves even lower token usage than Shortest 8/24 across all sections, except in AIME 25’s Solution phase, where Shortest 8/24 produces stronger reductions. Adaptive Difficulty GFPO also trims the Solution and Verification sections substantially, though less aggressively than the other two methods. Similar trends appear on GPQA.
\begin{findings}
GFPO slashes verbosity in the \textbf{solution} and \textbf{verification} phases of reasoning---cutting 94.4\% of excess length in the solution and 66.7\% of the excess length in the verification steps on AIME 25.
\end{findings}
The GFPO variants leave the Problem and Final segments largely unchanged (within 10\% of SFT counts), showing that GFPO specifically targets verbosity and redundancy in the core reasoning steps while preserving both the problem statement and the final answer.

%% file: figures/len_red_problem_diff_wrap.tex
\begin{figure}[t]
    \centering
    \begin{subfigure}[t]{0.49\textwidth}
        \centering
        \includegraphics[width=\textwidth, height=4.5cm]{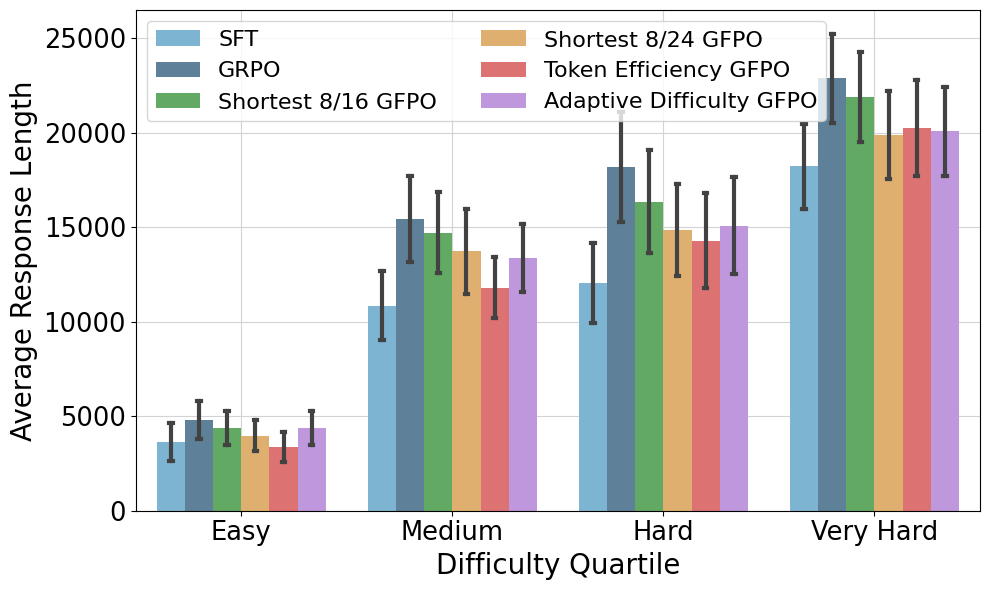}
        \caption{\centering \textbf{Average Response Length Across Problem Difficulties.}}
        \label{fig:len_over_diff}
    \end{subfigure}
    \hfill
    \begin{subfigure}[t]{0.49\textwidth}
        \centering
        \includegraphics[width=\textwidth, height=4.5cm]
        {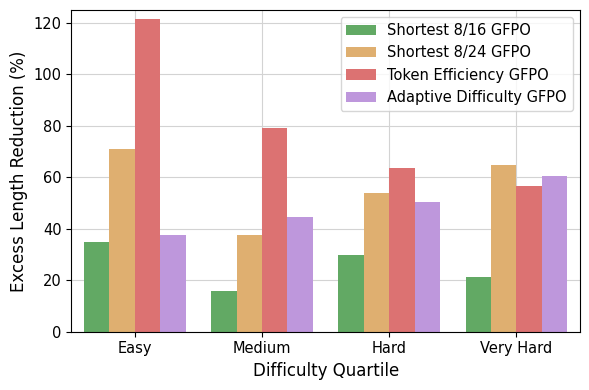}
        \centering
        \caption{\centering\textbf{ Excess Length Reduction Across Problem Difficulties.}}
        \label{fig:len_red_over_diff}
    \end{subfigure}
    \caption{\textbf{Average Length and Excess Length Reductions Across Problem Difficulties.} Response lengths rise with problem difficulty for all methods (\textbf{left}), but GFPO reduces excess length across all problem difficulty levels (\textbf{right}). Token efficiency has the most significant reductions--with responses more brief than even the SFT baseline on easy questions. Shortest 8/24 has the strongest reductions on very hard questions. }
    \label{fig:len_and_elr_over_diff}
\end{figure}

%% file: figures/analysis_second_figs_wrap.tex
\begin{figure}[t]
    \centering
    \begin{subfigure}[t]{0.49\textwidth}
        \centering
        \includegraphics[width=1.05\textwidth, height=5cm]{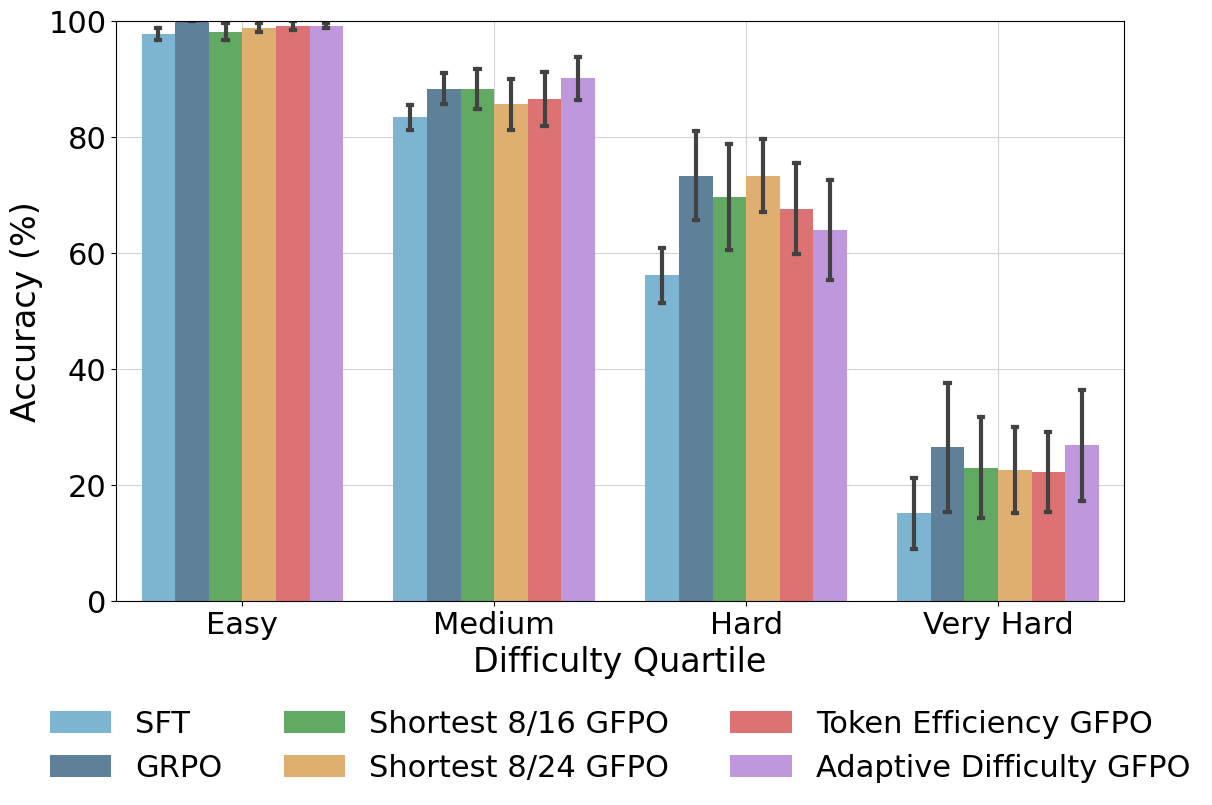}
        \caption{\textbf{Accuracy Across Problem Difficulties.} Adaptive Difficulty and Shortest 8/24 have the best accuracies. Token Efficiency has strongest length cuts, but with small non-statistically significant drops in accuracy.}
        \label{fig:acc_vs_diff_quartile}
    \end{subfigure}
    \hfill
    \begin{subfigure}[t]{0.49\textwidth}
        \centering
        \includegraphics[width=\textwidth, height=5cm]{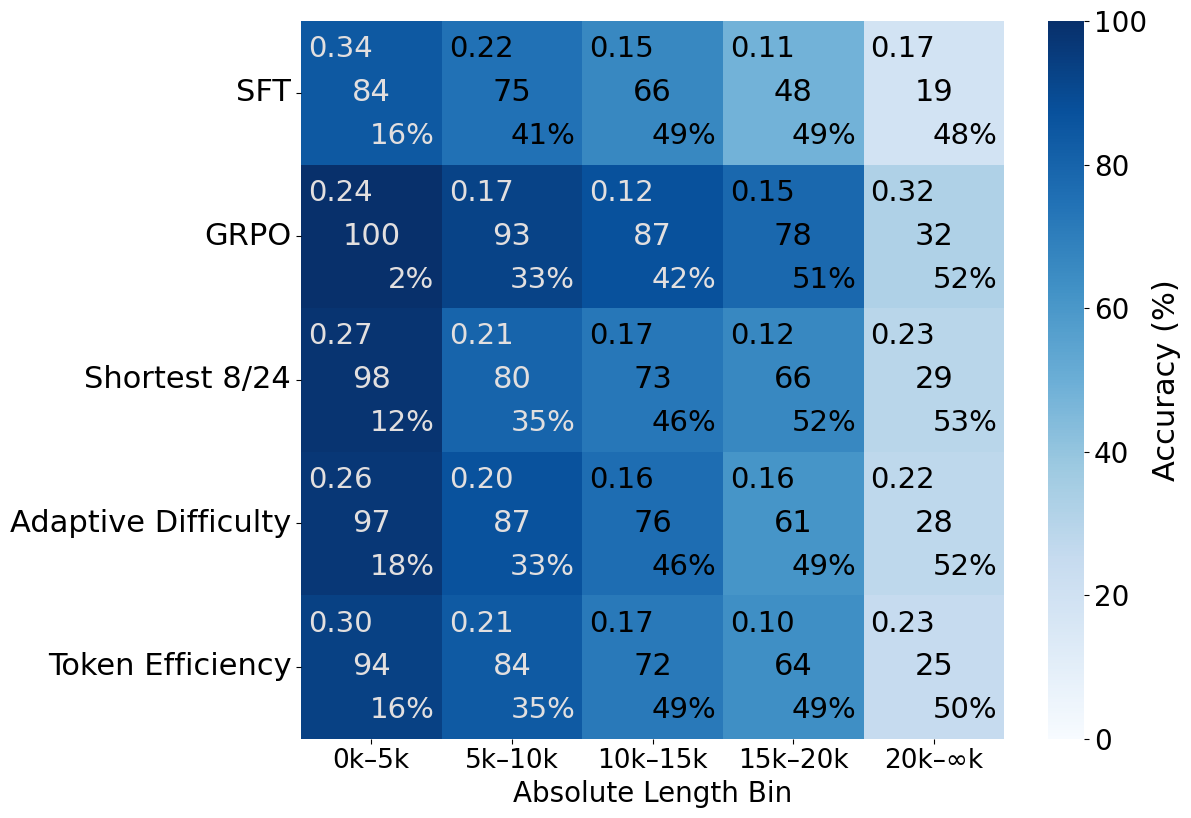}
        \caption{\textbf{Accuracy, Response Share, and Prompt Difficulty by Response Length.} Each cell shows accuracy (center), response share (top left), and prompt difficulty (bottom right; avg difficulty ($\leftarrow 1-SFT_{acc}$) of prompts corresponding to responses in cell, for a fixed response length range.}
        \label{fig:abs_len_heatmap}
    \end{subfigure}
    \caption{\textbf{GFPO Accuracy Across Difficulty Levels and Response Lengths.} \textbf{(a) Accuracy Across Problem Difficulties} shows Adaptive Difficulty matching or exceeding GRPO accuracy on easy, medium, and very hard questions with Shortest 8/24 matching GRPO on hard problems via larger $G$. \textbf{(b) Accuracy, Response Share, and Prompt Difficulty by Response Length} shows GFPO cuts long-tail verbosity (32\% to 22\% outputs $\geq20k$) and solves hard problems with shorter responses ($\sim$9x harder prompts solved with $\leq5k$ tokens).}
    \label{fig:acc_len_combined}
\end{figure}

%% file: figures/relative_len_lineplot_wrap.tex
\begin{wrapfigure}{r}{0.5\textwidth}
  \vspace{-0.7\baselineskip}
  \centering
  \includegraphics[width=0.5\textwidth]{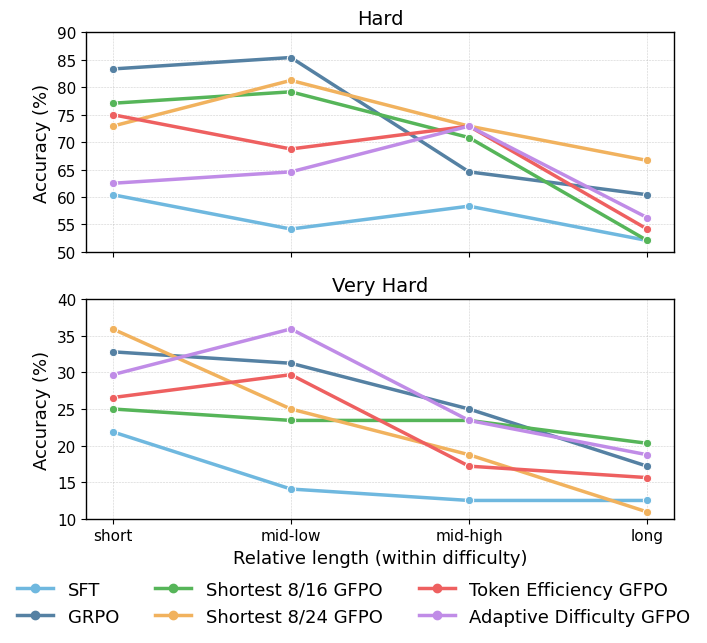}
  \caption{\textbf{Accuracy vs Relative Length for Hard and Very Hard Problems.} Accuracy declines with increasing response length even at fixed difficulty. On hard problems, most models peak at $12k-16k$ tokens, while GFPO variants outperform GRPO in the longest bin by producing shorter, more accurate long responses. On very hard problems, Adaptive Difficulty is most robust.
}
  \label{fig:acc_rel_len_hard}
  \vspace{-2.5\baselineskip}
\end{wrapfigure}

%% file: figures/len_kde_wrap.tex
\begin{wrapfigure}{l}{0.55\textwidth}
  \vspace{-0.7\baselineskip}
  \centering
  \includegraphics[width=\linewidth]{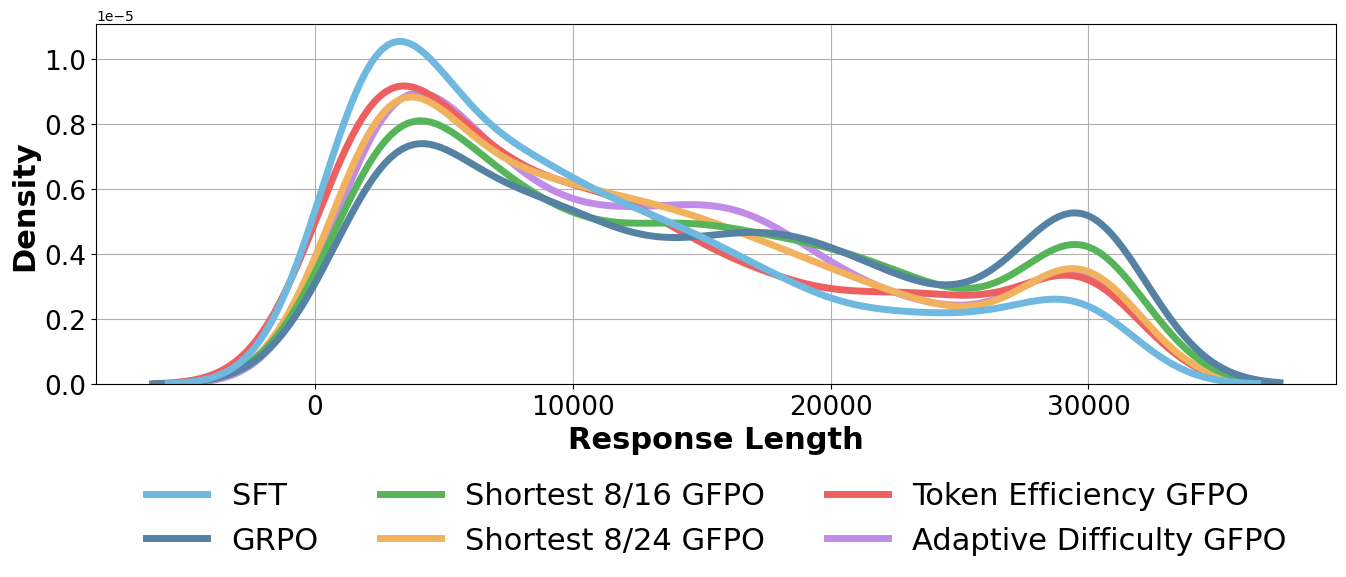}
  \caption{\textbf{Response Length Kernel Density.} Distribution of response lengths across all outputs on AIME 25. All GFPO variants compress the long tail  shifting distribution mass toward shorter outputs. Token Efficiency and Shortest 8/24 achieve the largest shifts.}
  \label{fig:len_kde}
  \vspace{-0.9\baselineskip}
\end{wrapfigure}

%% file: figures/acc_kde_wrap.tex
\begin{wrapfigure}{r}{0.55\textwidth}
  \vspace{-0.6\baselineskip}
  \centering
  \includegraphics[width=0.55\textwidth]{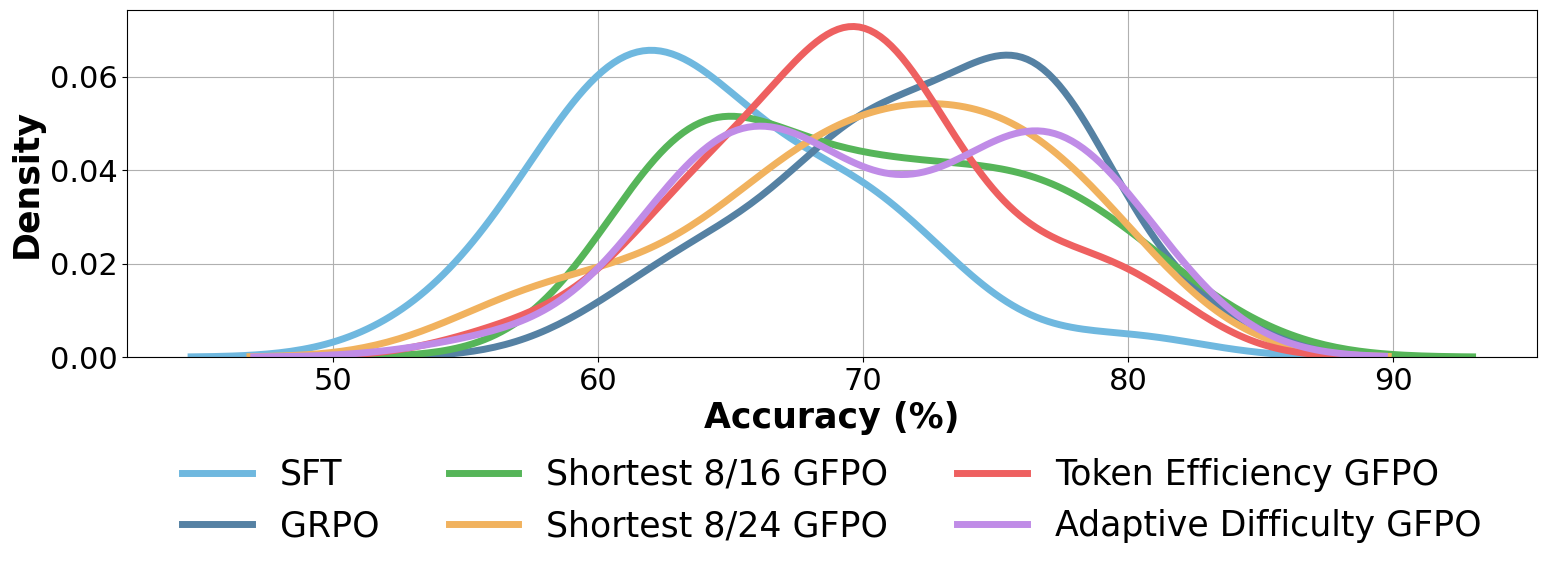}
  \caption{\textbf{Accuracy Kernel Density.} Per-repeat pass@1 accuracy distributions on AIME 25 from 32 independent runs for SFT, GRPO, and GFPO variant indicate the variance in performance of each model.}
  \label{fig:acc_kde}
  \vspace{-1.2\baselineskip}
\end{wrapfigure}

%% file: figures/token_analysis.tex
\begin{figure}[t]
    \centering
    \begin{minipage}[b]{0.48\textwidth}
        \includegraphics[width=\textwidth]{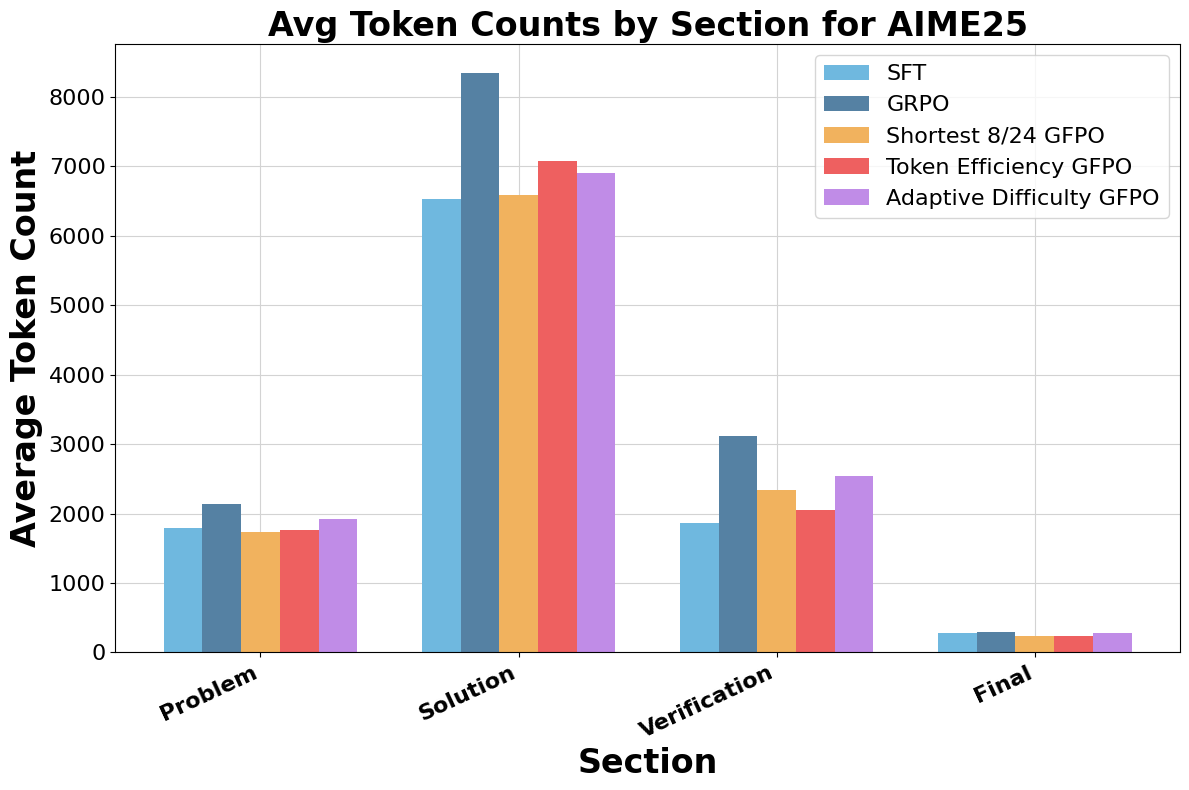}
    \end{minipage}
    \hfill
    \begin{minipage}[b]{0.48\textwidth}
        \includegraphics[width=\textwidth]{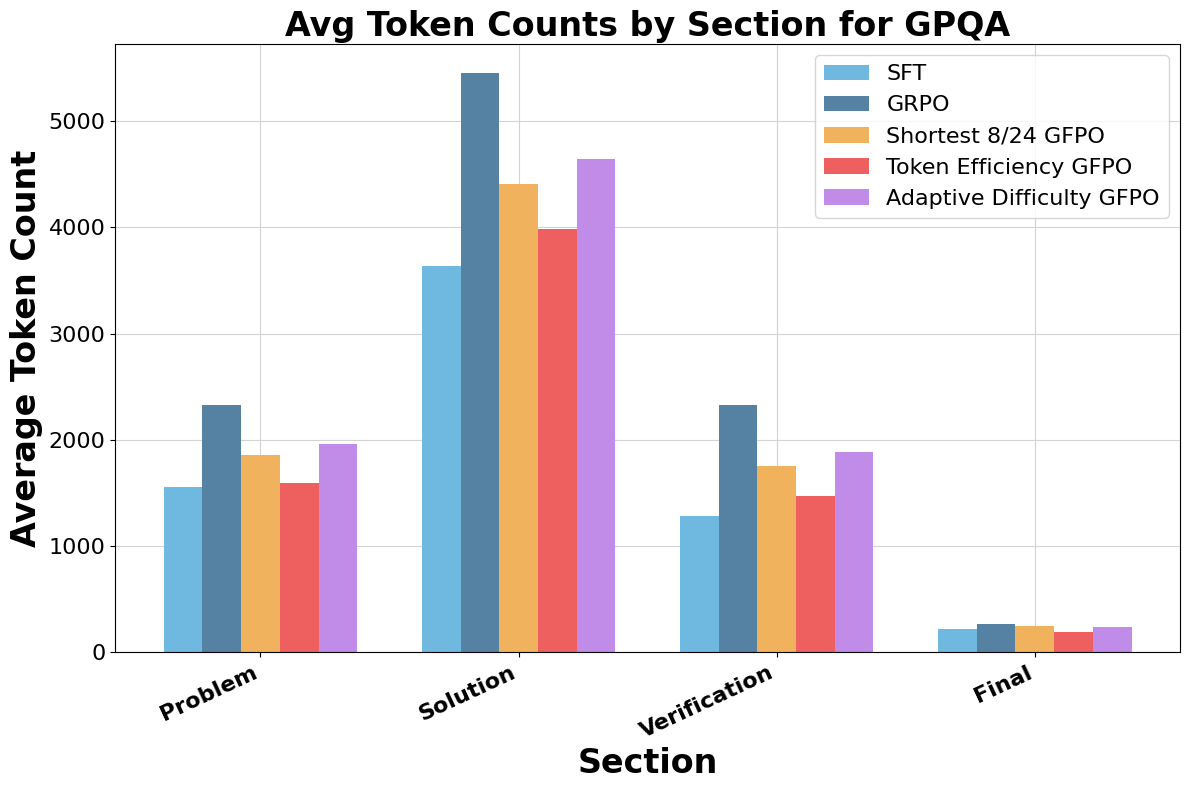}
    \end{minipage}
    \caption{\textbf{Average Token Counts by Reasoning-Trace Section for AIME 25 (left) and GPQA (right).} GRPO expands the Solution and Verification phases relative to SFT. GFPO variants markedly reduce this excess—on AIME 25, Shortest 8/24 cuts Solution length by 94.4\% and Verification length by 66.7\%.}
    \label{fig:token_analysis_aime_gpqa}
\end{figure}

%% file: figures/qual_example_main_body.tex
\vspace{-2\baselineskip}
\begin{AIbox}[breakable]{AIME 25 II Problem 1: GRPO v/s Token Efficiency GFPO}
\vspace{5mm}
\footnotesize

% -- "User" / <input> BOX (on top) --
\begin{tcolorbox}[
    colback=user!20,         % e.g. #FF8282 if you had \definecolor{user}{HTML}{FF8282}
    colframe=user!40!black,
    boxrule=0.5pt,
    arc=2pt,
    left=5pt,right=5pt,top=3pt,bottom=3pt,
    title={\bf User:}
]
Six points $A, B, C, D, E$ and $F$ lie in a straight line in that order. Suppose that $G$ is a point not on the line and that $AC = 26$, $BD = 22$, $CE = 31$, $DF = 33$, $AF = 73$, $CG = 40$, and $DG = 30$. Find the area of $\triangle BGE$.
\end{tcolorbox}
\begin{tcolorbox}[
    colback=grpo!50,        % background
    colframe=grpo!40!black, % border
    boxrule=0.5pt,
    arc=2pt,
    left=5pt,right=5pt,top=3pt,bottom=3pt,
    title=GRPO
]
\texttt{<think>}... 
So now triangle $BGE$: $B=(18,0)$, $G=(58,24)$, $E=(57,0)$. Its base $BE$ = distance from $18$ to $57$ = $39$. But then what is the height? The triangle's base is along x-axis and $G$ has y coordinate $24$, but note that $G$ is not vertically above the segment? Actually, base $BE$: endpoints $(18,0)$ and $(57,0)$. The distance from $G$ to line $BE$ is simply the absolute y coordinate if $G$'s x coordinate lies between $18$ and $57$? But here $x_G=58$, so $G$ is to the right of $E$. But the formula for area using base and height with base along line is the distance from $G$ to the line containing $BE$. But the line containing $BE$ is horizontal ($y=0$). So the distance is $|24| = 24$, regardless of the x-coordinate. So area $= \frac{1}{2} \cdot \text{base} \cdot \text{height} = \frac{1}{2} \cdot 39 \cdot 24 = 468$.
\\
...\\
Wait, $39 \cdot 24 = 936$, half is $468$. But wait, is that the answer? Let's check: base $BE$ length $= 57 - 18 = 39$, height $= 24$. So area $= \frac{39 \cdot 24}{2} = 468$.\\
...\\
But the problem's answer is expected to be an integer? Possibly yes. But let's check step 1: ``Six points $A, B, C, D, E, F$ lie in a straight line in that order." Our computed positions are consistent: $A=0$, $B=18$, $C=26$, $D=40$, $E=57$, $F=73$. Check distances: $AC=26$, $BD=40-18=22$, $CE=57-26=31$, $DF=73-40=33$, $AF=73$, correct. Now then triangles: $CG=$ distance from $C=(26,0)$ to $G=(58,24)$ equals $\sqrt{(32)^2+24^2}= \sqrt{1024+576}= \sqrt{1600}=40$, and $DG= \sqrt{(58-40)^2+24^2}= \sqrt{18^2+576}= \sqrt{324+576}= \sqrt{900}=30$, so correct. Now triangle $BGE$: vertices $B=(18,0)$, $G=(58,24)$, $E=(57,0)$. The base $BE$ has length $57-18=39$, and the height is the vertical distance from $G$ to the x-axis is $24$, so area$= \frac{39 \cdot 24}{2}=468$. So answer is $468$.\\
...\\
However, double-check: It might be that $G$ could have negative $y$. But then area is same magnitude, absolute value. Therefore answer: $468$.\\
...\\
But wait, check problem statement: ``Find the area of triangle $BGE$." Not necessarily with base $BE$ horizontal? But we computed that. I want to check if there's any trick: Possibly the triangle is not oriented with base along the line containing $B$ and $E$ because then the altitude would be the distance from $G$ to that line. And indeed the line $BE$ is horizontal, so the altitude is the absolute value of $G$'s $y$-coordinate. We computed $G$'s $y$-coordinate as $24$. So final answer: $468$.\\
...\\
Wait, but what about possibility $x=58$? Let's double-check the $x$-coordinate from difference of circle equations: We had circles: $(x -26)^2+y^2=40^2=1600, \ (x-40)^2+y^2=30^2=900$. Subtracting: $(x-26)^2 - (x-40)^2=700.$ Expand: $(x^2 -52x+676) - (x^2-80x+1600)= -52x+676+80x-1600=28x-924=700$, so $28x=1624, \ x=58$. So that's consistent. So yes. Thus final answer: $468$.
......
\texttt{</think>}
\end{tcolorbox}
% -- "Model2" / <reasoning><answer> BOX (below) --
\begin{tcolorbox}[
    colback=model!50,        % e.g. #BEE4D0 if you had \definecolor{model}{HTML}{BEE4D0}
    colframe=model!40!black,
    boxrule=0.5pt,
    arc=2pt,
    left=5pt,right=5pt,top=3pt,bottom=3pt,
    title=Token Efficiency GFPO
]
\texttt{<think>}...
Now we want the area of triangle $BGE$. The vertices are $B = (18, 0)$, $G = (58, 24)$ or $(58, -24)$, and $E = (57, 0)$. The area is given by $\frac{1}{2}$ absolute value of cross product of $\overrightarrow{BG}$ and $\overrightarrow{BE}$ etc. Let's compute vectors: $\overrightarrow{BG} = G - B = (58 - 18, 24 - 0) = (40, 24)$. $\overrightarrow{BE} = E - B = (57-18, 0-0) = (39, 0)$. Area $= \frac{1}{2}|\det(\overrightarrow{BG}, \overrightarrow{BE})| = \frac{1}{2} | (40 \cdot 0 - 24 \cdot 39) | = \frac{1}{2}| -936 | = 468$. Thus answer: $468$. So final answer: $468$.
......
\texttt{</think>}
\end{tcolorbox}

\begin{tcolorbox}[
    colback=comparison!20,         % e.g. #FF8282 if you had \definecolor{user}{HTML}{FF8282}
    colframe=comparison!40!black,
    boxrule=0.5pt,
    arc=2pt,
    left=5pt,right=5pt,top=3pt,bottom=3pt,
    title={\bf Comparison:}
]
The GRPO response recomputes the area of $\triangle BGE$ six times, while Token Efficiency GFPO computes it only once.
\end{tcolorbox}
\end{AIbox}

%% file: sections/related_work.tex
\section{Related Work}
\textbf{GRPO Loss Modifications.} 
Recent works such as Dr. GRPO~\citep{drgrpo} and DAPO~\citep{dapo} modify GRPO’s loss normalization to improve token efficiency and training stability. Standard GRPO normalizes loss within each response before averaging, giving all responses equal weight---resulting in the downweighting of tokens in longer outputs. Dr. GRPO instead normalizes by the maximum response length in the batch, and DAPO by the total token count—both increasing the weight of tokens in longer responses to more heavily penalize incorrect long chains.

Following open-source RL training frameworks such as \texttt{verl}~\citep{sheng2024hybridflow} and \texttt{TRL}~\citep{vonwerra2022trl}, GFPO uses DAPO’s token-level normalization. However, this not only penalizes long incorrect chains but also boosts the reward for long correct ones, often driving stronger SFTed reasoning models (e.g., Phi-4-reasoning-plus, DeepSeek-R1-Distill-Qwen) toward greater verbosity. This highlights the limits of loss normalization alone for length control. GFPO instead modifies the advantage function by computing advantages only for retained chains, a change independent of loss normalization; while our experiments pair it with DAPO’s loss aggregation, it could be combined with alternatives such as Dr. GRPO in future work.

\textbf{Length-Aware Penalties.} Beyond normalization, several works add explicit length-aware penalties to the reward in GRPO to discourage longer reasoning chains: \citet{hou2025thinkprune} impose a token-limit during RL (zero reward beyond the cap) and iteratively tighten it; \citet{su2025thinking} use an adaptive direct length penalty that evolves over training to curb over/under-compression; \citet{xiang2025just} scale the penalty inversely with per-prompt solve rate so easy prompts pay more for extra tokens; \citet{cheng2025optimizing} combine a global length reward with a targeted compress reward to remove redundant thinking; and \citet{aggarwal2025l1} optimize accuracy subject to a prompt-specified target length by penalizing deviations.
In our initial runs, simply scaling the length penalty in our length-aware reward did not yield substantial length reductions, or reduced length at the cost of accuracy. In contrast, GFPO’s rejection step \textit{implicitly} shapes the reward by determining which samples are used for learning, offering a simpler way to optimize for multiple properties (e.g., length, safety) without complex reward engineering. At the same time, pairing GFPO with a more carefully crafted reward could potentially deliver further gains.

\textbf{Inference-time Interventions.}  Other works have also explored how to control reasoning length purely at inference-time. Like our work, \citet{hassid2025don} show that shorter chains are often more accurate (even when controlling for hardness) and propose voting over the shortest $m$ of $k$ samples. \citet{muennighoff2025s1} introduce ``budget forcing'', where special phrases such as ``Wait'' or ``Final Answer'' are used to control when reasoning stops without re-training. Other approaches monitor intermediate generation signals and halt when the model appears confident or when the answer stabilizes across consecutive reasoning chunks \citep{liu2025answer, yang2025dynamic}. These methods are complementary to GFPO and could be combined to further reduce inference-time costs or enforce length constraints at post-training time.

\textbf{Rejection Sampling Methods.} 
Rejection sampling has been applied in various LLM training and decoding settings. \citet{kim2024m} explore post-training for length reduction \textit{after} RL by sampling multiple solutions per prompt and either (i) fine-tuning on the shortest correct response or (ii) applying DPO with the shortest correct output as a positive example and longer responses as negatives. In contrast, GFPO integrates rejection sampling \textit{within} the RL update—sampling a larger group per prompt and training on chains selected by length or reward-per-token—avoiding extra distillation, length-alignment stages, or explicit penalty terms.

DAPO~\citep{dapo} employs “dynamic sampling” discarding prompts whose responses are all correct or incorrect to stabilize batch gradients. \citet{xiong2025minimalist} show that a simple RAFT baseline \citep{dong2023raft}, which trains solely on positively rewarded samples  can perform comparably to GRPO. While their filtering is based on correctness reward, GFPO filters by length or token efficiency and can adaptively adjust the degree of filtering based on question difficulty.
Other applications include \citet{khaki2024rs} which combines rejection sampling with DPO by sampling more responses per prompt, and selecting contrastive pairs using reward scores;\citet{lipkin2025fast}, which uses adaptive weighted rejection sampling for efficient constrained generation; and \citet{sun2024fast}, which accelerates Best-of-$N$ decoding by early-rejecting low-scoring candidates with a reward model.

%% file: sections/conclusion.tex
\section{Conclusion} We introduce Group-Filtered Policy Optimization (GFPO), a simple yet versatile extension of GRPO that samples more responses and selectively learns from those best aligned with a desired attribute, such as brevity or token efficiency. This selective learning serves as implicit reward shaping, guiding policies toward higher-quality outputs while curbing undesirable behaviors like response length inflation without complex reward engineering. Across diverse reasoning benchmarks, length-optimized GFPO consistently preserves GRPO-level reasoning ability while substantially improving reasoning efficiency. Token Efficiency GFPO delivers the strongest length reductions overall, while Adaptive Difficulty GFPO applies test-time scaling only when needed—achieving the best efficiency-accuracy trade-offs on the hardest problems. By applying selective learning at training-time, GFPO converts a one-off increase in training-time compute into lasting gains in inference-time reasoning efficiency.

%% file: sections/acknowledgments.tex
\section{Acknowledgments}
We would like to thank Guoqing Zheng, Suriya Gunasekar, Piero Kauffmann, Mojan Javaheripi, Safoora Yousefi, and the AI Frontiers team at Microsoft Research for valuable discussions and feedback throughout this project.

%% file: sections/appendix.tex
\appendix

\section{Qualitative Examples}\label{appendix:qual_ex}
\input{figures/qualitative_examples_aime25}
\input{figures/qualitative_examples_aime25_p2}
\newpage
\subsection{GPQA Qualitative Examples.}
\input{figures/qualitative_examples_gpqa}

\section{Response Length and Accuracy for Hard and Very Hard Problems}
\input{tables/avg_len_acc_hard_len_bins}

%% file: figures/qualitative_examples_aime25.tex
\vspace{-0.7\baselineskip}
\begin{AIbox}[breakable]{AIME 25 I Problem 8: GRPO v/s Token Efficiency GFPO}
\vspace{5mm}
\footnotesize

% -- "User" / <input> BOX (on top) --
\begin{tcolorbox}[
    colback=user!20,         % e.g. #FF8282 if you had \definecolor{user}{HTML}{FF8282}
    colframe=user!40!black,
    boxrule=0.5pt,
    arc=2pt,
    left=6pt,right=6pt,top=4pt,bottom=4pt,
    title={\bf User:}
]
Let $k$ be real numbers such that the system $|25+20i-z|=5$ and $|z-4-k|=|z-3i-k|$ has exactly one complex solution $z$. The sum of all possible values of $k$ can be written as $\frac{m}{n}$, where $m$ and $n$ are relatively prime positive integers. Find $m+n$. Here $i=\sqrt{-1}$.
\end{tcolorbox}

% -- "Model" / <reasoning><answer> BOX (below) --
\begin{tcolorbox}[
    colback=grpo!50,        % background
    colframe=grpo!40!black, % 
    boxrule=0.5pt,
    arc=2pt,
    left=6pt,right=6pt,top=4pt,bottom=4pt,
    title=GRPO
]
\texttt{<think>}... 
Equate squares: $(x - (4+k))^2 + y^2 = (x-k)^2 + (y-3)^2$. Expand: $(x - (4+k))^2 = (x - k - 4)^2 = (x-k)^2 - 2 \cdot 4 \cdot (x-k) + 16$, wait check: actually, let's expand properly. Let $A = 4+k$, $B = k$, then equation becomes $(x - A)^2 + y^2 = (x - B)^2 + (y-3)^2$. Expand LHS: $(x - A)^2 = x^2 - 2A x + A^2$ and then plus $y^2$, so LHS $= x^2 -2A x + A^2 + y^2$. RHS: $(x - B)^2 + (y-3)^2 = (x^2 - 2B x + B^2) + (y^2 -6y+9) = x^2 - 2B x + B^2 + y^2 -6y +9$. Cancel $x^2$ and $y^2$ yields $-2A x + A^2 = -2B x + B^2 -6y+9$. Rearranging: $-2A x + A^2 + 2B x - B^2 + 6y - 9 = 0$, or $2(B-A)x + 6y + (A^2 - B^2 - 9) = 0$. But note that $A^2 - B^2 = (A+B)(A-B)$ and $A+B = (4+k)+k = 4+2k$, and $A-B = (4+k) - k = 4$, so $A^2 - B^2 = 4(4+2k) = 16+8k$. Then our equation is $2(B-A)x + 6y + (16+8k - 9) = 0$, i.e. $2(B-A)x + 6y + (7+8k) = 0$, but note that $B-A = k - (4+k) = -4$. Then $2(B-A)x = 2(-4)x = -8x$, so the equation becomes $-8x + 6y + 7+8k = 0$ or $8x - 6y = 7+8k$.\\
...\\
Alternatively, we can get the perpendicular bisector of the segment connecting $(4+k,0)$ and $(k,3)$. Let's check the midpoint: $M = \left(\frac{4+k+k}{2}, \frac{0+3}{2}\right) = \left(\frac{4+2k}{2}, \frac{3}{2}\right) = (2+ k, 3/2)$. The slope of the line joining these two points from $(4+k,0)$ to $(k,3)$ equals $\frac{3 - 0}{k - (4+k)} = \frac{3}{-4} = -\frac{3}{4}$. So the perpendicular bisector slope is the positive reciprocal: $\frac{4}{3}$, and it passes through $M$. Its equation is $y - 3/2 = \frac{4}{3}(x - (2+k))$. Multiply both sides by $3$: $3y - 9/2 = 4(x - 2 - k)$. Multiply both sides by $2$: $6y - 9 = 8(x - 2 - k) = 8x -16 - 8k$. Then rearranging: $6y - 9 = 8x - 16 - 8k$, so $8x - 6y = 16 + 8k + 9 = 25 + 8k$.\\
...\\
Let's re-check...
% actually, we need to recalc the equation from the distance formula.
% Let's do it carefully. 
% We want the set of points $z$ such that $|z - (4+k)| = |z - (3i+k)|$. Let $z = x + yi$, then the equation becomes: distance from $(4+k,0)$ to $(x,y)$ equals distance from $(k,3)$ to $(x,y)$.
So the equation is $(x - (4+k))^2 + (y-0)^2 = (x - k)^2 + (y - 3)^2$. Expand LHS: $(x - 4 - k)^2 + y^2 = (x - k - 4)^2 + y^2 = (x^2 - 2x(4+k) + (4+k)^2) + y^2$. The RHS is $(x - k)^2 + (y - 3)^2 = x^2 - 2kx + k^2 + y^2 - 6y + 9$. Cancel $x^2$ and $y^2$ from both sides to get $- 2(4+k)x + (4+k)^2 = -2kx + k^2 - 6y + 9$. Bring terms to one side: $- 2(4+k)x + (4+k)^2 + 2kx - k^2 + 6y - 9 = 0$. Combine the $x$ terms: $-2(4+k)x + 2kx = -8x - 2kx + 2kx = -8x$. Then the constant terms: $(4+k)^2 - k^2 - 9$; expanding $(4+k)^2 = k^2 + 8k + 16$, subtracting $k^2$ gives $8k+16$, and subtracting $9$ gives $8k+7$. So the equation becomes $-8x + 6y + (8k+7) = 0$, or equivalently $8x - 6y = 8k+7$.\\
......
\texttt{</think>}
\end{tcolorbox}

% -- "Model2" / <reasoning><answer> BOX (below) --
\begin{tcolorbox}[
    colback=model!50,        % e.g. #BEE4D0 if you had \definecolor{model}{HTML}{BEE4D0}
    colframe=model!40!black,
    boxrule=0.5pt,
    arc=2pt,
    left=6pt,right=6pt,top=4pt,bottom=4pt,
    title=Token Efficiency GFPO
]
\texttt{<think>}...
Squaring both sides yields: $(x - (4+k))^2 + y^2 = (x - k)^2 + (y - 3)^2$. Expand: $(x^2 -2(4+k)x + (4+k)^2 + y^2) = (x^2 -2kx + k^2 + y^2 - 6y + 9)$. Cancel $x^2$ and $y^2$, leaving: $-2(4+k)x + (4+k)^2 = -2kx + k^2 - 6y + 9$. Simplify: Bring terms: $-2(4+k)x + (4+k)^2 + 2kx - k^2 + 6y - 9 = 0$. Combine $x$ terms: $-2(4+k)x + 2kx = -2(4+k - k)x = -2 \cdot 4 x = -8x$. So equation becomes: $-8x + (4+k)^2 - k^2 + 6y - 9 = 0$. Simplify constant: $(4+k)^2 - k^2 = (16 + 8k + k^2) - k^2 = 16 + 8k$. So the equation is: $-8x + (16+8k) + 6y - 9 = 0$, or $-8x + 6y + (8k + 7) = 0$, but check: $16+8k - 9 = 7+8k$. Multiply by $-1$: $8x - 6y - (8k+7) = 0$, or in standard form: $8x - 6y = 8k+7$.
......
\texttt{</think>}
\end{tcolorbox}

\begin{tcolorbox}[
    colback=comparison!20,         % e.g. #FF8282 if you had \definecolor{user}{HTML}{FF8282}
    colframe=comparison!40!black,
    boxrule=0.5pt,
    arc=2pt,
    left=6pt,right=6pt,top=4pt,bottom=4pt,
    title={\bf Comparison:}
]
GRPO response recomputes and checks the perpendicular bisector equation 5 times (only 3 instances shown), while Token Efficiency GFPO computes it only once.
\end{tcolorbox}
\end{AIbox}

%% file: figures/qualitative_examples_aime25_p2.tex
\begin{AIbox}[breakable]{AIME 25 II Problem 5: GRPO v/s Shortest 8/24}
\vspace{5mm}
\footnotesize

% -- "User" / <input> BOX (on top) --
\begin{tcolorbox}[
    colback=user!20,         % e.g. #FF8282 if you had \definecolor{user}{HTML}{FF8282}
    colframe=user!40!black,
    boxrule=0.5pt,
    arc=2pt,
    left=6pt,right=6pt,top=4pt,bottom=4pt,
    title={\bf User:}
]
Suppose $\triangle ABC$ has angles $\angle BAC = 84^\circ$, $\angle ABC = 60^\circ$, and $\angle ACB = 36^\circ$. Let $D$, $E$, and $F$ be the midpoints of sides $\overline{BC}$, $\overline{AC}$, and $\overline{AB}$, respectively. The circumcircle of $\triangle DEF$ intersects $\overline{BD}$, $\overline{AE}$, and $\overline{AF}$ at points $G$, $H$, and $J$, respectively. The points $G$, $D$, $E$, $H$, $J$, and $F$ divide the circumcircle of $\triangle DEF$ into six minor arcs, as shown. Find $\wideparen{DE} + 2 \cdot \wideparen{HJ} + 3 \cdot \wideparen{FG}$, where the arcs are measured in degrees.

\begin{tikzpicture}[scale=1.2]
\coordinate (B) at (0,0);
\coordinate (C) at (6,0);
\coordinate (A) at (1.78,3.07);

\coordinate (D) at ($(B)!0.5!(C)$);
\coordinate (E) at ($(A)!0.5!(C)$);
\coordinate (F) at ($(A)!0.5!(B)$);

\draw (A) -- (B) -- (C) -- cycle;

\fill (A) circle (2pt) node[above] {$A$};
\fill (B) circle (2pt) node[below] {$B$};
\fill (C) circle (2pt) node[below] {$C$};

\fill (D) circle (2pt) node[below] {$D$};
\fill (E) circle (2pt) node[right] {$E$};
\fill (F) circle (2pt) node[left] {$F$};

\draw (D) -- (E) -- (F) -- cycle;

\coordinate (O) at (2.39,1.38);

\draw (O) circle (1.51);

\coordinate (G) at (1.72, 0.03);
\coordinate (J) at (1.5, 2.6);
\coordinate (H) at (2.08, 2.85);

\fill (G) circle (2pt) node[below] {$G$};
\fill (H) circle (2pt) node[above] {$H$};
\fill (J) circle (2pt) node[above left] {$J$};
\end{tikzpicture}
\end{tcolorbox}

% -- "Model" / <reasoning><answer> BOX (below) --
\begin{tcolorbox}[
    colback=grpo!50,        % background
    colframe=grpo!40!black, % 
    boxrule=0.5pt,
    arc=2pt,
    left=6pt,right=6pt,top=4pt,bottom=4pt,
    title=GRPO
]
\texttt{<think>} 
......
We need to compute arcs on the circumcircle of triangle $DEF$. Let's denote the vertices of triangle $DEF$ in order: It says circle of triangle $DEF$, and then the points $G, D, E, H, J, F$ are arranged in order along circle. So the vertices of triangle $DEF$ are $D, E, F$. But note the circle has six points in order, and the six arcs are (arc $DE$), then arc $EH$? Wait, but check the given: They mention "the points $G, D, E, H, J, \text{ and } F$ divide the circumcircle of triangle $DEF$ into six minor arcs, as shown." That means the circle order is: $G$, then $D$, then $E$, then $H$, then $J$, then $F$, and then back to $G$. But we need to be careful: The circumcircle of triangle $DEF$ has vertices $D, E, F$ as well. So the arcs are between adjacent points in the circle order. The six arcs are: arc $GD$, arc $DE$, arc $EH$, arc $HJ$, arc $JF$, arc $FG$. 
But the problem statement says "the points $G, D, E, H, J, \text{ and } F$ divide the circumcircle of triangle $DEF$ into six minor arcs, as shown." 
But then they ask "find arc $DE + 2 \cdot \text{arc } HJ + 3 \cdot \text{arc } FG$." But wait, which arc is arc $FG$? 
The arcs, I need to check which arcs are which from the order: They are given in the diagram: The points in order are: $G$, then $D$, then $E$, then $H$, then $J$, then $F$, then back to $G$. So the arcs are: arc $GD$, arc $DE$, arc $EH$, arc $HJ$, arc $JF$, arc $FG$. 
But the problem asks for arc $DE$, then $2 \cdot (\text{arc } HJ)$, then $3 \cdot (\text{arc } FG)$. But note that arc $FG$ is the arc from $F$ to $G$, but in our ordering, the arc from $F$ to $G$ is the arc $FG$, but careful: 
the order is $G$-$D$-$E$-$H$-$J$-$F$, so arc $FG$ is the arc from $F$ to $G$, which is indeed the last arc in the sequence. 
So we need to compute these arc measures, but they are not trivial. 
We need to find angles related to triangle $ABC$ and midpoints $D, E, F$. Let coordinates: $A, B, C$. But maybe we can use properties: The circle through midpoints of sides of a triangle is the medial circle, but it's not necessarily the circumcircle of the medial triangle, but that's exactly what it is: The circumcircle of triangle $DEF$, where $D, E, F$ are midpoints of $BC, AC, AB$ respectively, is the nine-point circle of triangle $ABC$.
.....
\texttt{</think>}
\end{tcolorbox}

% -- "Model2" / <reasoning><answer> BOX (below) --
\begin{tcolorbox}[
    colback=model!50,        % e.g. #BEE4D0 if you had \definecolor{model}{HTML}{BEE4D0}
    colframe=model!40!black,
    boxrule=0.5pt,
    arc=2pt,
    left=6pt,right=6pt,top=4pt,bottom=4pt,
    title=Shortest 8/24 GFPO
]
\texttt{<think>}
......
We need to compute arcs on the circumcircle of triangle $DEF$. Let's denote the vertices of triangle $DEF$ in order: It says circle of triangle $DEF$, and then the points $G, D, E, H, J, F$ are arranged in order along circle. So the vertices of triangle $DEF$ are $D, E, F$. But note the circle has intersections with lines $BD, AE, AF$. Let's re-read the problem statement. Given triangle $ABC$ with angles $A=84$, $B=60$, $C=36$. $D, E, F$ are midpoints of sides $BC, AC, AB$ respectively. The circumcircle of triangle $DEF$ intersects $BD, AE, AF$ at points $G, H, \text{ and } J$ respectively. The points $G, D, E, H, J, F$ divide the circumcircle of triangle $DEF$ into six minor arcs, as shown. I want to label positions: We have triangle $ABC$, but then midpoints $D, E, F$ are on sides $BC, AC, AB$ respectively. Now, note that triangle $DEF$ is the medial triangle of $ABC$. And the circumcircle of the medial triangle is known to be the nine-point circle of triangle $ABC$.
....
\texttt{</think>}
\end{tcolorbox}
\begin{tcolorbox}[
    colback=comparison!20,         % e.g. #FF8282 if you had \definecolor{user}{HTML}{FF8282}
    colframe=comparison!40!black,
    boxrule=0.5pt,
    arc=2pt,
    left=6pt,right=6pt,top=4pt,bottom=4pt,
    title={\bf Comparison:}
]
The Shortest 8/24 GFPO response quickly recognizes that $\triangle DEF$ is the medial triangle and its circumcircle is the nine-point circle, while the GRPO response gets bogged down in arc bookkeeping before arriving at the same conclusion.
\end{tcolorbox}
\end{AIbox}

\begin{AIbox}[breakable]{AIME 25 I Problem 9: GRPO v/s Adaptive Difficulty GFPO}
\vspace{5mm}
\footnotesize

% -- "User" / <input> BOX (on top) --
\begin{tcolorbox}[
    colback=user!20,         % e.g. #FF8282 if you had \definecolor{user}{HTML}{FF8282}
    colframe=user!40!black,
    boxrule=0.5pt,
    arc=2pt,
    left=6pt,right=6pt,top=4pt,bottom=4pt,
    title={\bf User:}
]
The parabola with equation $y = x^2 -4$ is rotated $60$ degrees counterclockwise around the origin. The unique point in the fourth quadrant where the original parabola and its image intersect has $y$-coordinate $\frac{a - \sqrt{b}}{c}$. Find $a+b+c$.
\end{tcolorbox}

% -- "Model" / <reasoning><answer> BOX (below) --
\begin{tcolorbox}[
    colback=grpo!50,        % background
    colframe=grpo!40!black, % 
    boxrule=0.5pt,
    arc=2pt,
    left=6pt,right=6pt,top=4pt,bottom=4pt,
    title=GRPO
]
\texttt{<think>} 
.....

But before that, we can attempt to factor the quartic if possible. It is a quartic in $u$. But perhaps a substitution might simplify it? Alternatively, note that $u$ might be a "nice" value? Alternatively, we can try to find the intersection by considering symmetry. Alternatively, consider the fact that the intersection point is the unique one in the fourth quadrant. Maybe we can try to solve directly by eliminating $u$. Alternatively, consider using rotation in reverse: For an intersection point $(x,y)$, we have $y = x^2 - 4$ and also $(y + 4)$??? Not sure.
\\
......
\texttt{</think>}
\end{tcolorbox}

% -- "Model2" / <reasoning><answer> BOX (below) --
\begin{tcolorbox}[
    colback=model!50,        % e.g. #BEE4D0 if you had \definecolor{model}{HTML}{BEE4D0}
    colframe=model!40!black,
    boxrule=0.5pt,
    arc=2pt,
    left=6pt,right=6pt,top=4pt,bottom=4pt,
    title=Adaptive Difficulty GFPO
]
\texttt{<think>} 
.....

I think we need to solve for $x$ explicitly from the quartic. Alternatively, we can use the approach: Let $(x,y)$ be the intersection point. Then $(x,y)$ satisfies: $y = x^2 - 4$, and also $(x,y)$ satisfies the equation of the rotated parabola, i.e. if we rotate $(x,y)$ by $-60$ degrees, we get a point on the original parabola: So let $(u,v) = \left( \frac{1}{2}x + \frac{3}{2}y,\ - \frac{3}{2}x + \frac{1}{2}y \right)$. Then $y = u^2 - 4$.

......
\texttt{</think>}
\end{tcolorbox}
\begin{tcolorbox}[
    colback=comparison!20,         % e.g. #FF8282 if you had \definecolor{user}{HTML}{FF8282}
    colframe=comparison!40!black,
    boxrule=0.5pt,
    arc=2pt,
    left=6pt,right=6pt,top=4pt,bottom=4pt,
    title={\bf Comparison:}
]
The GRPO response jumps between several different alternatives---factoring the quartic, trying substitutions, symmetry arguments, and vague rotation ideas without committing to a plan and doesn't reduce uncertainty, whereas the Adaptive Difficulty GFPO defines a single clear approach to take.
\end{tcolorbox}
\end{AIbox}

%% file: figures/qualitative_examples_gpqa.tex
% \vspace{-0\baselineskip}
\begin{AIbox}[breakable]{GPQA: GRPO v/s Token Efficiency GFPO}
\vspace{5mm}
\footnotesize
% -- "User" / <input> BOX (on top) --
\begin{tcolorbox}[
    colback=user!20,         % e.g. #FF8282 if you had \definecolor{user}{HTML}{FF8282}
    colframe=user!40!black,
    boxrule=0.5pt,
    arc=2pt,
    left=6pt,right=6pt,top=4pt,bottom=4pt,
    title={\bf User:}
]
Consider this annihilation process $p+\bar{p}\rightarrow2A^{+}+2A^{-}$\\The antiproton is slowly moving and $m_{A}c^{2}=300MeV$.\\What is the velocity of particle A?
\end{tcolorbox}

% -- "Model" / <reasoning><answer> BOX (below) --
\begin{tcolorbox}[
    colback=grpo!50,        % background
    colframe=grpo!40!black, % 
    boxrule=0.5pt,
    arc=2pt,
    left=6pt,right=6pt,top=4pt,bottom=4pt,
    title=GRPO
]
\texttt{<think>}... 
Actually, maybe the process is in the lab frame where the antiproton is slow relative to the proton. But wait, which is slowly moving: "The antiproton is slowly moving" means that maybe the proton is moving? Wait check: "$p + \bar{p}$" one of them is slow, but which one? "The antiproton is slowly moving" means that the proton is moving fast? But then the question states: "Consider this annihilation process: $p + \bar{p} \to 2A^{+} + 2A^{-}$". The wording "The antiproton is slowly moving" implies that the initial momentum is coming from the proton, which is fast. But then the question is asking: "What is the velocity of particle $A$?" Possibly meaning the outgoing particles have velocity determined by energy-momentum conservation in the center-of-mass frame.\\
...\\
Alternatively, the problem might be set in the center-of-mass frame. Let me think: "Annihilation process" usually is considered in the center-of-mass frame. But then "The antiproton is slowly moving" might be a hint that the proton is at rest? Because usually one would take $p$ is at rest, and the anti-proton is slow relative to $p$? Actually, "slowly moving" means small velocity.  
... But if the antiproton is slow, then maybe the proton has kinetic energy? But then the process might be not exactly symmetric. Alternatively, maybe the assumption is that the antiproton is slow relative to the lab frame. That means the proton is at rest? Because then the antiproton is slow? That is contradictory: if the proton is at rest, then the anti-proton must be moving if we are in the lab frame? \\
...\\
Maybe the intended scenario is that the annihilation takes place in the lab frame where the antiproton is almost at rest, and the proton is moving with a velocity such that the center-of-mass energy is just above threshold. But then threshold for production of $4A$ is $4 m_A = 1200\ \mathrm{MeV}$.\\
...\\
That doesn't work. Maybe the intended scenario is that both the proton and antiproton are nearly at rest in the center-of-mass frame...\\
......
\texttt{</think>}
\end{tcolorbox}

% -- "Model2" / <reasoning><answer> BOX (below) --
\begin{tcolorbox}[
    colback=model!50,        % e.g. #BEE4D0 if you had \definecolor{model}{HTML}{BEE4D0}
    colframe=model!40!black,
    boxrule=0.5pt,
    arc=2pt,
    left=6pt,right=6pt,top=4pt,bottom=4pt,
    title=Token Efficiency GFPO
]
\texttt{<think>}...
Possibly the antiproton is slowly moving, so the reaction is in the lab frame, and one particle is at rest, but then the total kinetic energy available is the difference of the rest masses? But careful: if the antiproton is "slowly moving" that means it is non-relativistic? But then the proton might be at rest or antiproton at rest? Let me re-read: "The antiproton is slowly moving." So perhaps the proton is at rest? Actually, which one is at rest? The phrase "slowly moving" might indicate that the antiproton is nearly at rest relative to the proton? But then the total energy available is the mass difference, but then momentum conservation forces the final products to share momentum such that they have equal velocity magnitude. But which velocity? Let me analyze the reaction in the center-of-mass frame.
......
\texttt{</think>}
\end{tcolorbox}

\begin{tcolorbox}[
    colback=comparison!20,         % e.g. #FF8282 if you had \definecolor{user}{HTML}{FF8282}
    colframe=comparison!40!black,
    boxrule=0.5pt,
    arc=2pt,
    left=6pt,right=6pt,top=4pt,bottom=4pt,
    title={\bf Comparison:}
]
The GRPO response repeatedly rehashes whether the reaction is in the lab frame or center-of-mass frame, while Token Efficiency GFPO response correctly makes the assumption of center-of-mass frame once.
\end{tcolorbox}

\end{AIbox}
\begin{AIbox}[breakable]{GPQA: GRPO v/s Adaptive Difficulty GFPO}
\vspace{5mm}
\footnotesize
% -- "User" / <input> BOX (on top) --
\begin{tcolorbox}[
    colback=user!20,         % e.g. #FF8282 if you had \definecolor{user}{HTML}{FF8282}
    colframe=user!40!black,
    boxrule=0.5pt,
    arc=2pt,
    left=6pt,right=6pt,top=4pt,bottom=4pt,
    title={\bf User:}
]
A quantum mechanical particle of mass $m$ moves in two dimensions in the following potential, as a function of $(r,\theta)$: $V(r, \theta) = \frac{1}{2} k r^2 + \frac{3}{2} k r^2 \cos^2(\theta)$. Find the energy spectrum.
\end{tcolorbox}

% -- "Model" / <reasoning><answer> BOX (below) --
\begin{tcolorbox}[
    colback=grpo!50,        % background
    colframe=grpo!40!black, % 
    boxrule=0.5pt,
    arc=2pt,
    left=6pt,right=6pt,top=4pt,bottom=4pt,
    title=GRPO
]
\texttt{<think>}... 
Let $x = r \cos\theta$, $y = r \sin\theta$. Then $r^2 = x^2+y^2$, $\cos^2\theta = \frac{x^2}{r^2}$ if $r \neq 0$. So potential becomes: $V(x,y)= \frac{1}{2}k (x^2+y^2) + \frac{3}{2}k (x^2+y^2)\left(\frac{x^2}{x^2+y^2}\right) = \frac{1}{2}k (x^2+y^2) + \frac{3}{2}k x^2 = \frac{1}{2}k x^2 + \frac{1}{2}k y^2 + \frac{3}{2}k x^2 = (2)k x^2$? Let's check: $\frac{1}{2}k x^2 + \frac{3}{2}k x^2 = 2k x^2$, plus $\frac{1}{2}k y^2$ remains. So potential becomes: $V(x,y)= 2 k x^2 + \frac{1}{2}k y^2$. That is a two-dimensional harmonic oscillator with different frequencies along $x$ and $y$. But wait, does that match? Let's recalculate: Given $V(r,\theta)= \frac{1}{2} k r^2 + \frac{3}{2} k r^2 \cos^2\theta = \frac{1}{2} k (x^2+y^2) + \frac{3}{2} k \cos^2\theta (x^2+y^2)$. But $\cos\theta = x/r$ so $\cos^2\theta = \frac{x^2}{x^2+y^2}$. So then $V = \frac{1}{2} k (x^2+y^2) + \frac{3}{2} k (x^2+y^2) \cdot \frac{x^2}{x^2+y^2} = \frac{1}{2} k (x^2+y^2) + \frac{3}{2} k x^2$. So indeed $V = \frac{1}{2}k y^2 + \left(\frac{1}{2} + \frac{3}{2}\right)k x^2 = \frac{1}{2}k y^2 + 2k x^2$. That is a harmonic oscillator with frequencies: $\omega_x = \sqrt{\frac{4k}{m}} = 2\sqrt{\frac{k}{m}}$ and $\omega_y = \sqrt{\frac{k}{m}}$.\\
...\\
Wait, but our derivation assumed that the potential is separable in Cartesian coordinates, which is valid if we perform transformation $x = r \cos\theta$, $y = r \sin\theta$. But is that always valid? Let's double-check: The given potential is $V(r,\theta)= \frac{1}{2} k r^2 + \frac{3}{2} k r^2 \cos^2\theta = \frac{1}{2} k (x^2+y^2) + \frac{3}{2} k \left(\frac{x^2}{x^2+y^2}\right) (x^2+y^2) = \frac{1}{2}k (x^2+y^2) + \frac{3}{2} k x^2 = \frac{1}{2} k y^2 + 2k x^2$. Yes that is indeed a separable potential in Cartesian coordinates with $x$ and $y$ degrees of freedom. So the energy eigenvalues are those of two independent harmonic oscillators with frequencies $\omega_x = \sqrt{\frac{4k}{m}} = 2 \sqrt{\frac{k}{m}}$ and $\omega_y = \sqrt{\frac{k}{m}}$. \\
......
\texttt{</think>}
\end{tcolorbox}

% -- "Model2" / <reasoning><answer> BOX (below) --
\begin{tcolorbox}[
    colback=model!50,        % e.g. #BEE4D0 if you had \definecolor{model}{HTML}{BEE4D0}
    colframe=model!40!black,
    boxrule=0.5pt,
    arc=2pt,
    left=6pt,right=6pt,top=4pt,bottom=4pt,
    title=Adaptive Difficulty GFPO
]
\texttt{<think>}...
Let $x = r \cos\theta$, $y = r \sin\theta$. Then $\cos^2\theta = \frac{x^2}{r^2}$ if $r \neq 0$. So then potential becomes: $V(x,y)= \frac{1}{2} k (x^2+y^2) + \frac{3}{2}k (x^2+y^2) \left(\frac{x^2}{x^2+y^2}\right) = \frac{1}{2} k (x^2+y^2) + \frac{3}{2}k x^2$. So rewriting: $V(x,y)= \frac{1}{2} k y^2 + \left(\frac{1}{2} k + \frac{3}{2}k\right)x^2 = \frac{1}{2} k y^2 + 2k x^2$. Wait, check: $x^2 + y^2$ multiplied by factor $\frac{1}{2} k$ gives $\frac{1}{2} k x^2 + \frac{1}{2} k y^2$ and then add $\frac{3}{2}k x^2$ gives total potential: $V(x,y)= \left(\frac{1}{2} k+ \frac{3}{2}k\right)x^2 + \frac{1}{2} k y^2 = (2k)x^2 + \frac{1}{2}k y^2$. So then the potential is anisotropic harmonic oscillator in Cartesian coordinates with frequencies: for $x$: $\omega_x = \sqrt{\frac{4k}{m}}= 2 \sqrt{\frac{k}{m}}$ and for $y$: $\omega_y = \sqrt{\frac{k}{m}}$.\\
......
\texttt{</think>}
\end{tcolorbox}

\begin{tcolorbox}[
    colback=comparison!20,         % e.g. #FF8282 if you had \definecolor{user}{HTML}{FF8282}
    colframe=comparison!40!black,
    boxrule=0.5pt,
    arc=2pt,
    left=6pt,right=6pt,top=4pt,bottom=4pt,
    title={\bf Comparison:}
]
The GRPO response recalculates the harmonic oscillator frequencies multiple times while the Adaptive Difficulty GFPO response only calculates this once.
\end{tcolorbox}
\end{AIbox}
\newpage
\begin{AIbox}[breakable]{GPQA: GRPO v/s Shortest 8/24 GFPO}
\vspace{5mm}
\footnotesize
% -- "User" / <input> BOX (on top) --
\begin{tcolorbox}[
    colback=user!20,         % e.g. #FF8282 if you had \definecolor{user}{HTML}{FF8282}
    colframe=user!40!black,
    boxrule=0.5pt,
    arc=2pt,
    left=6pt,right=6pt,top=4pt,bottom=4pt,
    title={\bf User:}
]
A light beam is propagating through a glass with index of refraction n. The glass is moving at constant velocity v in the same direction as the beam and toward the observer in laboratory. What is the speed of light in glass relative to the observer in laboratory? Take the speed of light in vacuum $c=1$.
\end{tcolorbox}

% -- "Model" / <reasoning><answer> BOX (below) --
\begin{tcolorbox}[
    colback=grpo!50,        % background
    colframe=grpo!40!black, % 
    boxrule=0.5pt,
    arc=2pt,
    left=6pt,right=6pt,top=4pt,bottom=4pt,
    title=GRPO
]
\texttt{<think>}... 
Let me derive from special relativity: The speed of light in a medium is $c/n$ in the medium's rest frame. But when the medium is moving with velocity $v$, the speed of light in the lab frame is given by the relativistic velocity addition formula: $u' = \frac{u + v}{1+ (u v)/c^2}$. Here, $u = c/n$ and $v$ is the velocity of the medium in lab frame. But wait, is it $v$? But careful: The medium is moving with velocity $v$ in lab frame and the light moves in the medium frame with speed $c/n$. But the medium's velocity relative to lab is $v$, so we can use velocity addition. But then the effective speed in lab frame for light traveling in the same direction as medium is: $u = \frac{c/n + v}{1 + (v c/(n c^2))}$? Wait, let me recalc: The formula is: $u = \frac{u' + v}{1 + u' v/c^2}$. But $c=1$, so $u = \frac{1/n + v}{1 + v/(n)}$. Actually, check: If $u' = 1/n$ then $u = \frac{1/n + v}{(n+v)/n} = \frac{1/n + v}{(n+v)/n} = \left(\frac{1}{n} + v\right)\frac{n}{n+v} = \frac{1+ n v}{n+v}$. But wait, then answer is $\frac{1+n v}{n+v}$?\\
...\\
But is that the Fizeau drag effect? The Fizeau drag coefficient is $1 - 1/n^2$. Wait, I recall the formula for the speed of light in a moving medium in the direction of flow is: $u = 1/n + (1 - 1/n^2) v$. Let me check with our expression: $\frac{1/n + v}{1+ v/n} = \frac{1/n + v}{(n+v)/n} = \left(\frac{1}{n} + v\right)\frac{n}{n+v} = \frac{1+ n v}{n+v}$. Now expand that in $v$: Let $v$ be small. Write $\frac{1+ n v}{n+v} = \frac{1+ n v}{n(1+v/n)} = \frac{1}{n}\frac{1+ n v}{1+v/n}$. Now expand denominator: $\frac{1}{1+v/n}= 1 - v/n + (v/n)^2 + \dots$ So we get $= \frac{1}{n}(1+ n v)(1 - v/n) = \frac{1}{n}[1 + n v - v/n - n v \cdot (v/n)] = \frac{1}{n}[1 + n v - v/n - v^2]$. Actually, let me do it properly: $(1+ n v)(1 - v/n) = 1 - v/n + n v - v^2$. So that equals $1 + n v - v/n - v^2$. Dividing by $n$ gives $= 1/n + v - v/n^2 - v^2/n$. So the coefficient of $v$ is $v(1 - 1/n^2)$. So indeed, to first order in $v$, $u = 1/n + (1-1/n^2) v$. That is the Fizeau drag formula. So our expression is consistent with the known formula for the speed of light in a moving medium. But wait, are we using the correct sign? The medium is moving toward the observer, so the effective speed of light relative to the observer is increased relative to the medium's rest frame speed. So the answer is $u_{\mathrm{lab}} = \frac{1/n + v}{1+v/n}$ which simplifies to $\frac{1+n v}{n+v}$. \\
......
\texttt{</think>}
\end{tcolorbox}

% -- "Model2" / <reasoning><answer> BOX (below) --
\begin{tcolorbox}[
    colback=model!50,        % e.g. #BEE4D0 if you had \definecolor{model}{HTML}{BEE4D0}
    colframe=model!40!black,
    boxrule=0.5pt,
    arc=2pt,
    left=6pt,right=6pt,top=4pt,bottom=4pt,
    title=Shortest 8/24 GFPO
]
\texttt{<think>}...\\
The transformation from glass frame to lab frame, where the speed of light is $c/n$ in the glass frame and then add the velocity of glass, which is $v$. But note that the direction is the same as the beam. But the lab observer sees the beam speed as $v + (1/n)$? But wait not exactly, because the addition formula is: $u' = \frac{u+v}{1+u v}$. But careful: Actually, we have an object moving at speed $u$ (with respect to the glass) and the glass moving at speed $v$ (with respect to lab) along same direction. The lab speed $u_{\mathrm{lab}}$ of the light beam should be computed using relativistic addition formula: $u_{\mathrm{lab}} = \frac{u_{\mathrm{glass}} + v}{1+ u_{\mathrm{glass}} \cdot v}$. But note that $u_{\mathrm{glass}}$ is $1/n$. So lab speed $= \frac{1/n + v}{1 + (1/n)v}$. Multiply numerator and denominator by $n$: $= \frac{1 + n v}{n + v}$...But in lab frame, the medium is moving, so the speed of light in lab frame is: $u = \frac{1/n + v}{1 + v/n} = \frac{1+nv}{n+v}$ as computed. Wait check: Multiply numerator and denominator by $n$, get $\frac{1+nv}{n+v}$. Yes.
\\
......
\texttt{</think>}
\end{tcolorbox}

\begin{tcolorbox}[
    colback=comparison!20,         % e.g. #FF8282 if you had \definecolor{user}{HTML}{FF8282}
    colframe=comparison!40!black,
    boxrule=0.5pt,
    arc=2pt,
    left=6pt,right=6pt,top=4pt,bottom=4pt,
    title={\bf Comparison:}
]
The GRPO response circles back and re-derives the velocity addition with $u_\text{glass}=1/n$ multiple times, while the Shortest 8/24 GFPO response does this once with a small recheck.
\end{tcolorbox}
\end{AIbox}

%% file: tables/avg_len_acc_hard_len_bins.tex
\begin{table}[h]
\centering
\begin{tabular}{@{}ll|llll@{}}
\toprule
\textbf{Difficulty Bin} & \multicolumn{1}{c}{\textbf{Method}} & \multicolumn{1}{c}{\textbf{Short}} & \multicolumn{1}{c}{\textbf{Mid-Low}} & \multicolumn{1}{c}{\textbf{Mid-High}} & \multicolumn{1}{c}{\textbf{Long}} \\ \midrule
\rowcolor[HTML]{CFCFCF}\multicolumn{1}{l|}{Hard} & SFT & 7298 & 9949 & 12576 & 18349 \\
\rowcolor[HTML]{CFCFCF}\multicolumn{1}{l|}{Hard} & GRPO & 12719 & 16292 & 19846 & 23834 \\ \midrule
\multicolumn{1}{l|}{Hard} & Shortest 8/16 & 11292 & 13948 & 17897 & 22211 \\
\multicolumn{1}{l|}{Hard} & Shortest 8/24 & 10087 & 12839 & 15711 & 20837 \\
\multicolumn{1}{l|}{Hard} & Token Efficiency & \textbf{8918} & \textbf{12044} & \textbf{15337} & \textbf{20815} \\
\multicolumn{1}{l|}{Hard} & Adaptive Difficulty & 9593 & 12959 & 16126 & 21677 \\ \midrule
\rowcolor[HTML]{CFCFCF}\multicolumn{1}{l|}{Very Hard} & SFT & 10707 & 15630 & 20875 & 25666 \\
\rowcolor[HTML]{CFCFCF}\multicolumn{1}{l|}{Very Hard} & GRPO & 16728 & 22026 & 25309 & 27462 \\ \midrule
\multicolumn{1}{l|}{Very Hard} & Shortest 8/16 & 15768 & 20786 & 24051 & 26935 \\
\multicolumn{1}{l|}{Very Hard} & Shortest 8/24 & \textbf{12657} & \textbf{18219} & 22671 & \textbf{25911} \\
\multicolumn{1}{l|}{Very Hard} & Token Efficiency & 13034 & 18633 & 23223 & 26109 \\
\multicolumn{1}{l|}{Very Hard} & Adaptive Difficulty & 13096 & 18625 & \textbf{22276} & 26279 \\ \bottomrule
\end{tabular}
\caption{\textbf{Average Response Length by Difficulty and Length Bins.} We bin each model's responses to hard and very hard problems into length quartiles (short, mid-low, mid-high, long) and report the average response lengths across length bins. We \textbf{highlight} the shortest average response length per response length quartile across the different RL methods.}
\label{tab:app_len}
\vspace{2\baselineskip}
\begin{tabular}{@{}llllll@{}}
\toprule
\multicolumn{1}{c}{\textbf{Difficulty Bin}} & \multicolumn{1}{c}{\textbf{Method}} & \multicolumn{1}{c}{\textbf{Short}} & \multicolumn{1}{c}{\textbf{Mid-Low}} & \multicolumn{1}{c}{\textbf{Mid-High}} & \multicolumn{1}{c}{\textbf{Long}} \\ \midrule
\rowcolor[HTML]{CFCFCF}\multicolumn{1}{l|}{Hard} & \multicolumn{1}{l|}{SFT} & 60.42 & 54.17 & 58.33 & 52.08 \\
\rowcolor[HTML]{CFCFCF}\multicolumn{1}{l|}{Hard} & \multicolumn{1}{l|}{GRPO} & \textbf{83.33} & \textbf{85.42} & 64.58 & 60.42 \\ \midrule
\multicolumn{1}{l|}{Hard} & \multicolumn{1}{l|}{Shortest 8/16} & 77.08 & 79.17 & 70.83 & 52.08 \\
\multicolumn{1}{l|}{Hard} & \multicolumn{1}{l|}{Shortest 8/24} & 72.92 & 81.25 & \textbf{72.92} & \textbf{66.67} \\
\multicolumn{1}{l|}{Hard} & \multicolumn{1}{l|}{Token Efficiency} & 75.00 & 68.75 & \textbf{72.92} & 54.17 \\
\multicolumn{1}{l|}{Hard} & \multicolumn{1}{l|}{Adaptive Difficulty} & 62.50 & 64.58 & \textbf{72.92} & 56.25 \\ \midrule
\rowcolor[HTML]{CFCFCF}\multicolumn{1}{l|}{Very Hard} & \multicolumn{1}{l|}{SFT} & 21.88 & 14.06 & 12.50 & 12.50 \\
\rowcolor[HTML]{CFCFCF}\multicolumn{1}{l|}{Very Hard} & \multicolumn{1}{l|}{GRPO} & 32.81 & 31.25 & \textbf{25.00} & 17.19 \\ \midrule
\multicolumn{1}{l|}{Very Hard} & \multicolumn{1}{l|}{Shortest 8/16} & 25.00 & 23.44 & 23.44 & \textbf{20.31} \\
\multicolumn{1}{l|}{Very Hard} & \multicolumn{1}{l|}{Shortest 8/24} & \textbf{35.94} & 25.00 & 18.75 & 10.94 \\
\multicolumn{1}{l|}{Very Hard} & \multicolumn{1}{l|}{Token Efficiency} & 26.56 & 29.69 & 17.19 & 15.63 \\
\multicolumn{1}{l|}{Very Hard} & \multicolumn{1}{l|}{Adaptive Difficulty} & 29.69 & \textbf{35.94} & 23.44 & 18.75 \\ \bottomrule
\end{tabular}
\caption{\textbf{Accuracy (\%) by Difficulty and Length Bins.} We bin each model's responses to hard and very hard problems into length quartiles (short, mid-low, mid-high, long) and report the accuracies across length bins. We \textbf{highlight} the highest accuracy per response length quartile across the different RL methods.}
\label{tab:app_acc}
\end{table}